\documentclass[journal]{IEEEtran}
\usepackage[table]{xcolor}
\usepackage{graphicx}
\usepackage[tight,footnotesize]{subfigure}
\usepackage{adjustbox}
\usepackage[T1]{fontenc}
\usepackage{rotating}
\usepackage{amsmath} 
\usepackage{booktabs}

\ifCLASSINFOpdf

\else
 
\fi

\begin{document}
%
\title{Decoding Islamophobic Discourse: Using LLMs to Identify Tropes and Semi-Coded Hate Speech}
%
%
%

\author{Raza Ul Mustafa,
        Roi Dupart,
       Gabrielle Smith, 
       Noman Ashraf,
       Nathalie Japkowicz
\thanks{Raza Ul Mustafa, Roi Dupart are with Loyola University New Orleans, USA e-mail: rulmust@loyno.edu, radupart@my.loyno.edu}
\thanks{Gabrielle Smith is from Tulane University, New Orleans, USA, e-mail (gsmith33@tulane.edu).}

\thanks{Noman Ashraf is an Independent researcher, e-mail (nomanashraf712@gmail.com).}
\thanks{Nathalie Japkowicz is from American University, USA, e-mail (japkowic@american.edu).}

}

%
%

\markboth{Journal of \LaTeX\ Class Files,~Vol.~14, No.~8, August~2015}%
{Shell \MakeLowercase{\textit{et al.}}: Bare Demo of IEEEtran.cls for IEEE Journals}
%



\maketitle

\begin{abstract}
In recent years, Islamophobia\footnote{{\color{red} Warning: This paper contains examples of racist and Islamophobic/Antisemitic statements that may be disturbing to the reader.}} has gained significant traction across Western societies, fueled by the rise of digital communication networks. This paper performs a large-scale analysis of specialized, semi-coded Islamophobic terms such as (muzrat, pislam, mudslime, mohammedan, muzzies) floated on extremist social platforms, i.e., 4Chan, Gab, Telegram, etc. Many of these terms appear lexically neutral or ambiguous outside of specific contexts, making them difficult for both human moderators and automated systems to reliably identify as hate speech. First, we use Large Language Models (LLMs) to show their ability to understand these terms. Second, we use Google Perspective API, which suggests that Islamophobic posts tend to receive higher toxicity scores than other categories of hate speech, like Antisemitism. Finally, we show the capabilities of LLMs in predicting tropes expressed in Islamophobic discourse. Our findings indicate that LLMs understand Out-Of-Vocabulary (OOV) and semi-coded slurs; however, further improvements in moderation strategies and algorithmic detection are needed to address this discourse effectively. Moreover, LLMs continue to exhibit limitations in understanding tropes in Islamophobia. Taken together, we conducted one of the first large-scale studies of semantically coded and OOV terms and shed light on Islamophobia globally.
\end{abstract}

\begin{IEEEkeywords}
hate speech, Islam, Islamophobia, social media, Antisemitism, GPT, OpenAI
\end{IEEEkeywords}

%
\IEEEpeerreviewmaketitle

\begin{table*}
\centering
\caption{Semi-coded (OOV) terms and their use on extremist social platforms.}
\begin{tabular}{|p{12cm}|p{2.5cm}|}
\hline
Text                                                                                                                                                               & Terms      \\ \hline
\parbox{12cm}{\vspace{0.1cm}{\color{red}Mohammedans} are incompatible with Western Society!!! {\color{red}Mohammedans} need to be sent back to their {\color{red}Mohammedan} shitholes!!!\vspace{0.1cm}}           & Mohammedan \\ \hline
\parbox{12cm}{\vspace{0.1cm}The {\color{red}mudslimes} of the Indian subcontinent are the most subhuman.  Arab/MENA Muslims are very civilized in comparison.\vspace{0.1cm}}              & Mudslime   \\ \hline
\parbox{12cm}{\vspace{0.1cm}i'd rather take a ride on the challenger than fly with a {\color{red}muzrat} as pilot, lmao. \vspace{0.1cm}}                                                                                             & Muzrat \cite{awan2016islamophobia}     \\ \hline
\parbox{12cm}{\vspace{0.1cm}Can’t wait till Europeans go full Reconquista and force every {\color{red}pislam}\\ worshipper to convert or get deported.\vspace{0.1cm}}                         & Pislam     \\ \hline
\parbox{12cm}{\vspace{0.1cm}{\color{red}muzzies} outbreed whites it doesn't matter they'll just make more {\color{red}mudslimes}\vspace{0.1cm}  }                                                                                                 & Muzzies    \\ \hline
\end{tabular}
\label{tab:ex-terms}
\end{table*}
\section{Introduction}
\IEEEPARstart{T}here has been a significant increase in hate speech on social media platforms, e.g., X, 4Chan, Gab, etc. In the past, offensive language was used on these platforms to incite Islamophobia \cite{mohideen2008language,qu2023unsafe}; however, the use of coded, semi-coded and out-of-vocabulary (OOV) language is currently experiencing a significant surge. 

\textit{We define coded language or terms as ambiguous, presenting itself as neutral while embedding hateful meanings. Here, we use \textit{neutrality} to refer to lexical ambiguity; these words are not considered offensive according to social standards when evaluated without additional context.} 
Similarly, we define hate speech as direct and serious attacks on any protected category of people based on their race, religion, ethnicity, national origin, sex, or gender. We adopted this definition from prior literature \cite{hine2017kek, davidson2017automated} and standards defined in social networks \cite{x_hateful_conduct_policy, meta_hateful_conduct_policy}.
%
Such terms may appear harmless or ordinary at first glance, making them less detectable for Large Language Models (LLMs). Some are real words, such as \textit{Abdul}, which is simply an Arabic word or a common Muslim name; others are altered versions of real words, like \textit{muzzie.} These words are constructed intentionally to evade detection, often by mimicking neutrality or through slight spelling changes. For example, \textit{mudslime} is formed by inserting the words \textit{mud} and \textit{slime} into the original word \textit{Muslim.} While a human reader may eventually recognize the intended target, the prefix and suffix mislead both readers and algorithms, making the hateful intent less obvious, especially to LLMs trained on standard language usage. The deliberate use of coded, ambiguous, or distorted words presents a significant challenge for automated hate speech detection. To better explain coded and OOV words, the following post was posted on the Bitchute social media platform on August 10, 2024.
\begin{itemize}
    \item ``muzrats are gonna be getting it any day now. pack your bags abdul'' -- Bitchute (2024-08-10)
\end{itemize}

In the above post, it is clear that an Islamophobic connotation is implied. Muzz is a British Muslim marriage and dating app. However, \textit{muzrats} are two words, muz – muzz and rat(s) are various medium-sized, long-tailed rodents. Similarly, \textit{abdul} is also used in an Islamophobic connotation, a neutral but hateful slur against Muslims. To add more context, use of \textit{abdul} is highlighted below in both coded and benign form.

\begin{itemize}
    \item ``slow your roll   \colorbox{red}{abdul}. we all hate kik*s, paj**t, sikhs/hindus, chinks, spics, turk roaches   and sand ni**ers. take your mosques and madrasas to yurope.'' - 4Chan -- 30-06-2025
    \item ``go marry your cousin, \colorbox{red}{abdul}.'' - 4Chan -- 17-06-2025
    \item ``angel of mercy: remembering pakistan‚ \colorbox{green}{abdul} sattar edhi, nine years on edhi‚legacy continues to inspire acts of compassion and service across the globe'' - Telegram (09-07-2025) -- benign
\end{itemize}

 \textit{Abdul Sattar Edhi} is a humanitarian, philanthropist, and ascetic who runs the Edhi foundation in Pakistan. 


Similarly, in the post below posted on 4Chan, dated November 5, 2024, on the US election day, 2024, the Islamophobic connotation is quite clear; however, the words tank and thank both have different meanings. A tank is an armored fighting vehicle intended as a primary offensive weapon in front-line ground combat, whereas thank is an expression of gratitude. Moreover, by adding context to the message, \textit{abdul} is an Islamic name; however, it is a neutral word used in both ways. 
\begin{itemize}
    \item ``Your idiotism won us elections, tank you \colorbox{red}{abdul}. Now get lost.''
\end{itemize}

Similarly, on extremist platforms, we find other terms like \textit{mohammedan}, \textit{pislam}, \textit{shitskin} and \textit{muzzies}. These terms are used derogatorily to imply something negative about the Muslim identity, differentiating and alienating Muslims in a subtle but damaging way. Thus, awareness of these terminologies, similar to \textit{abdul} and OOV words such as \textit{pislam}, \textit{muzrats}, \textit{mohammedan}, is necessary for accurate hate speech detection against Muslims. 
Recent advancements in Natural Language Processing (NLP), including LLMs (e.g., GPT, Llama), provide a contextual understanding of text \cite{mihalcea2024developments}. We also argue that LLMs like GPT-4 and recent versions understand these OOV terminologies (slur) used against Muslims. Therefore, we recommend that social platforms integrate these OOV and coded terms and related expressions into their content moderation strategies to create a more inclusive and respectful online environment. We believe that such slurs are used against Muslims to blame them for various issues worldwide.  
Therefore, in this research, we try to verify the following hypotheses: 

\begin{itemize}
    \item \textbf{H1:} We hypothesize that advanced LLMs understand the OOV words used against Muslims. Moreover, in recent years, the most floated trope on extremist social media platforms is \textit{Muslim immigration to white countries}.
    \item \textbf{H2:} Hate speech against Muslims is more violent with toxic content, such as the use of abusive and racial remarks.   
    \item \textbf{H3:} Islamophobic discourse on these extremist social platforms is characterized by various tropes that advanced LLMs struggle to distinguish, despite their capabilities. 
\end{itemize}

To verify our hypothesis \textbf{H1}, we use GPT-4 to classify OOV and neutral terms. We selected a random sample from our labeled dataset, which includes these terms. For hypothesis \textbf{H2}, we compare the toxicity of Antisemitism with Islamophobia. We select Antisemitism because it targets Jewish people based on their religious, ethnic, or cultural identity; in this respect, it is similar to Islamophobia.
For hypothesis \textbf{H3}, we first manually annotated the dataset using expert knowledge and then used LLMs capabilities to predict tropes expressed in the discourse.

The rest of the paper is organized as follows: In Section \ref{tab:background}, we provide background and state-of-the-art work, followed by a description of the dataset collection for our study in Section \ref{tab:dataset_labeling}. Next, Section \ref{tab:results} discusses the results of all the hypotheses. We conclude our paper in Section \ref{tab:conclusion} and discuss future work directions.

\section{Background and State-of-the-art}\label{tab:background}

Hate speech often employs slurs and stereotypes to demean race, religion, or cultural practices \cite{howard2017freedom,zsisku2024hate, watanabe2018hate,burnap2015cyber}. Attacks on race, for instance, may generalize all Muslims as belonging to specific racial groups, such as Arabs, South Asians, or Africans, using derogatory terms rooted in colonial or racial hierarchies. Similarly, religious hate speech shifts focus to Islam as a belief system, often mocking its practices or associating it with violence or extremism. Likewise, anti-Muslim rhetoric has recently surged, especially directed at Muslim immigrants in Western nations, where racial tensions are already high. This rhetoric often includes xenophobic language, dehumanizing Muslims as \textit{creatures} or accusing them of \textit{stealing} a country. 

Another recurring theme in Islamophobia is to target the Prophet Muhammad, which frequently involves defamatory claims designed to discredit him as a religious figure. Common tropes include references to his multiple wives, participation in wars, or accusations about the ages of his wives. For instance, accusing the Prophet of being a \textit{pedophile} or \textit{rapist}, attempting to delegitimize him through inflammatory language. A phrase like \textit{piss be upon him} mocks the Islamic phrase \textit{``peace be upon him''}, aiming to insult the Prophet and, by extension, the faith he represents. Finally, there are stereotypes about Muslim women, particularly surrounding modesty practices, polygamy, and cultural traditions.

These religious themes often contain OOV and neutral terms such as \textit{muzrat, pislam, mudslime, mohammedan and muzzies,} to discuss various political tensions and target Muslims. These terms are introduced in a way to convey a disrespectful and toxic message to Muslims. We show an example of each OOV term in Table \ref{tab:ex-terms}.\footnote{In  Antisemitism, a commonly known hate-glossary is available online --  https://www.ajc.org/translatehateglossary. We took some help from the hate glossary of Islamophobia at - https://hatebase.org/search\_results. }

\subsection{Islamophobia Derogatory Terms} \label{tab:dterms}
\noindent \textbf{Mudslime and Muzzies:}
The term \textit{mudslime} is a deliberate distortion of the word muslim, using negative prefixes and connotations to vilify followers of Islam. The prefix \textit{mud-} suggests filth or something undesirable, while \textit{slime} further degrades the term by evoking imagery of something repulsive or unclean. Despite its altered spelling, \textit{mudslime} is phonetically similar to Muslim, indicating an intentional effort to avoid saying the proper name of the religion while still making it recognizable enough to serve as an insult. The same is used for \textit{muzzies}. This type of linguistic manipulation strips Muslims of their religious identity by replacing it with a derogatory imitation.

\noindent \textbf{Pislam:} 
Similar to \textit{mudslime} and \textit{muzzies}, the term \textit{pislam} distorts the word \textit{Islam} by inserting a negative prefix. The \textit{pis-} prefix, derived from \textit{piss} connotes waste or impurity. This deliberate modification strips Islam of its dignity and respect, replacing it with a term meant to provoke disgust. Like other slurs that alter the pronunciation of \textit{Islam} or Muslim, this one reflects a broader pattern of linguistic dehumanization, where opponents refuse to acknowledge the proper names of Islamic faith and identity, instead opting for insults that reduce them to mockery.

\noindent \textbf{Mohammedan and Muzrat} 
\textit{Mohammed} is simply a name, while the \textit{dan} suffix comes from Arabic, representing a vowel sound known as \textit{tanween}, often used to indicate a grammatical form or relationship in Arabic. It’s unclear whether this derogatory term originated from the Arabic version with \textit{tanween} or if the \textit{dan} part was explicitly created by hate speech users with an additional meaning. Regardless, \textit{mohammadan} has a derogatory intent. The use of this term reflects hate speech that targets the Prophet Muhammad, a central figure in Islam. To call someone a \textit{mohammadan} implies they are a follower of Prophet Muhammad, and because the term is used by those who view the Prophet negatively, it becomes an insult to Muslims, as it attacks a respected religious figure. 

The term \textit{muzrat}~\cite{awan2016islamophobia} is similar to other slurs like \textit{mudslime} and \textit{muzzies}, but it differs in that it doesn't even attempt to sound like the word \textit{Muslim}. While terms like \textit{mudslime} distort Muslim with negative prefixes, \textit{muzrat} takes a more direct approach to mocking. By adding the word \textit{rat} at the end, the term reduces Muslims to a vermin-like status, implying that they are unwanted or subhuman. This form of mockery is more straightforward in its intent: it strips Muslims of their identity and uses degrading language to associate them with something repulsive.

\subsection{State of the Art}
Research on hate speech consistently shows that religious groups, particularly Muslims, are among the most frequent targets, with Islamophobia appearing as the dominant category in several large-scale analyses \cite{Vidgen2019TrajectoriesOI}, \cite{Elsherief2018HateLA}. Moreover, Islamophobic implicit discourse is often driven not only by prejudice but by a combination of factors such as political ideology, anti-immigration attitudes, misinformation, and economic anxieties; therefore, highlighting the dynamic nature of both perpetrators and the hate speech itself \cite{Vidgen2019TrajectoriesOI}. 

Hate speech in any form against any minority has documented harms for both individuals and the community, weakening feelings of safety, belonging, and social inclusion. Continuous exposure to hate speech reinforces negative stereotypes and contributes to increased societal polarization. Since most Islamophobia hate speech is online, its impact is often underestimated by observers who are not direct targets \cite{Obermaier2021IllBT}. Rosen et al. also suggest that hate speech can damage not only individuals but also the overall quality of online discussions by conducting a large-scale analysis of hate speech on Reddit~\cite{rosen2025antisemitic}. They found that when a hateful comment appears, replies become more limited, since people repeat similar words and ideas rather than introduce new ones. Therefore, hate speech doesn’t just hurt users of the online digital space; it also makes discussions smaller and less creative.


Bleich \cite{Bleich2011WhatII} defines Islamophobia as \textit{indiscriminate negative attitudes or emotions directed at Islam or Muslims,} emphasizing that prejudice is a multidimensional phenomenon involving explicit, implicit, overt, and covert elements. This complexity reveals limitations in many computational approaches. For instance, tools like the Perspective API classify toxicity using categories such as insult, profanity, identity attack, threat, and overall toxicity \cite{Weber2025DigitalGC}. However, because prejudice can take implicit forms, it is inappropriate to frame hate speech as a binary category, as done in several earlier works \cite{Ahn2024SharedConIH}, \cite{Calderon2021LinguisticPF}, and \cite{Gitari2015ALA}. Vidgen et.al, \cite{Vidgen2018DetectingWA}, attempt to address this by adding a \textit{weak hate} category, while \cite{Talat2016HatefulSO} argues that collapsing all abusive behavior into one label obscures important distinctions and advocates for more fine-grained classifications. Previous studies \cite{aldreabi2023enhancing}, \cite{GonzlezBaquero2023TheCA}, \cite{Saeed2023TopicMB} also show recurring themes in Islamophobic discourse that align with well-documented tropes \cite{Mustafa2024CanGD} as follows.  


\begin{itemize}
    \item \textbf{Women:} women, burqa, hijab, oppress, marry
    \item \textbf{Immigration:} leave, nation, country, western
    \item \textbf{Religion:} cult, Mohammed, radical, violent
    \item \textbf{Violence:} kill, rape
    \item \textbf{Race:} negro, black
    \item \textbf{Global conflict/terrorism:} Taliban, jihadi, Israel, terrorist
\end{itemize}
While these studies identify thematic clusters, none classify them into tropes. We argue that tropes are often loaded with political tensions and misinformation, which fuels the use of coded and OOV language against Muslims. Therefore, this work's central focus is on coded and OOV language and the state-of-the-art LLMs' capabilities for understanding them, rather than offensive-language categorization.

%
The authors in \cite{aldreabi2023enhancing}, explore the use of deep learning models to detect Islamophobia on social media by applying two distinct definitions and measuring toxicity levels. Although the article demonstrates promising results in identifying overtly Islamophobic language, this success is largely due to the reliance on easily flaggable keywords like \textit{terrorist} or \textit{hate} that can easily be flagged as offensive language \cite{jaleel2023islamophobia}. The authors in \cite{farrand2023hate} argue that simple binary classifications of Islamophobia are inadequate for capturing the complexity of hate speech on social media. Their study uses a dataset of 109,488 tweets from far-right Twitter accounts collected in 2017, classifying content as non-Islamophobic, weak Islamophobia, and strong Islamophobia. Strong Islamophobia includes direct derogatory statements or calls for discrimination, while weak Islamophobia comprises subtler, culturally coded remarks, such as critiques of specific events or cultural practices. However, the definitions of Islamophobia don’t delve into distinct types of hate that Muslims experience. There are many layers to Islamophobia beyond simple negativity or hostility, such as religious hostility, political discrimination, and cultural stereotyping.


Authors in \cite{mozafari2020hate} present a study that leverages a BERT-based model to explore hate speech detection while mitigating bias, particularly concerning racial and dialectal language. This study does not fully cover implicit hate speech and there's no focus on training the model to be adaptable in detecting hate speech across other culturally significant dialects or contexts without further refinement. Similarly, in another study, the author used a graph neural network for Islamophobia \cite{wasi2024explainable}. They build a graph where nodes represent text and are connected by a similarity matrix. By this, the author classified the text as Islamophobic instead of relying on individual words in a post. They tried various combinations of embeddings; however, the GNN + BERT-based embedding significantly outperformed previous methods (accuracy ~ 0.75, macro-F1 ~ 0.747), demonstrating that the graph-based approach performs very well on Islamophobia. However, the work is an explicit detection of hate speech, and we focused more on implicit and LLMs capabilities for both Islamophobia and Antisemitism. Building on prior research, another work by~\cite{boyuk2024hate} uses explicit expressions such as \textit{terrorist}, \textit{jihad}, \textit{sharia}, and \textit{Islamist} to collect data from TikTok. The authors find that the TikTok algorithm quickly spreads such content, which is responsible for spreading negativity and hate crimes against Muslims.

In the context of LLMs, Roy, Sarthak, et al. used three datasets: HateXplain, Implicit Hate, and ToxicSpans on GPT-3.5, text‑davinci‑003, and Flan‑T5‑large for hate speech and toxic language classification without giving any additional information in a zero-shot manner \cite{roy2023probing}. They also analyze model performance by varying prompt strategies, such as how hate speech is defined and the targeted community. They find that, instead of zero-shot prompting, models given additional information about the community and the type of hate achieve accuracy improvements of ~20–30\%. 

Since hate speech is a multilingual problem, authors investigate LLMs capabilities for hate speech detection across languages such as Spanish, Portuguese, German, French, Italian, Turkish, Hindi, and Arabic \cite{ghorbanpour2025can}. Compared to zero-shot in \cite{roy2023probing}, the authors used both zero-shot and few-shot prompting and compared the results with traditional encoder models. They find that prompting requires different styles across languages and that these differences greatly influence the results. 
Authors in \cite{sen2024hatetinyllm} provided HateTinyLLM, a fine-tuned model that uses tiny-sized decoder-only large language models (tiny LLMs), such as TinyLlama (1.1B parameters), Phi-2, and OPT-1.3B. They used the technique, LoRA (Low-Rank Adaptation), and adapter layers on two datasets DynaHate and HateEval. They also highlight the importance of fine-tuning tiny-LLMs on hate-speech data rather than using computationally expensive LLMs. Similarly, in another work, Kikkisetti et al. \cite{kikkisetti2024coded} fine-tuned a BERT model for coded-term discovery. In their work, they used widely known antisemitism terms~\footnote{https://www.ajc.org/translatehateglossary} and collected data from Pyrra. This work is quite close to research in this regard, as we also worked on coded language; however, in Islamophobia, most of the terms are semi-coded and OOV. For instance, consider an example of the coded term \textit{control} in Antisemitism compared to the semi-coded terms such as \textit{Mohammedan, Mudslime}  in Islamophobia. Here, control can be used in both ways, toxic and benign, in Antisemitism, whereas Islamophobic terms can't be used the same way.

Compared to literature, we primarily rely on LLMs capabilities, such as the state-of-the-art GPT-4, to classify OOV terms and tropes.
These OOV terms, which are used derogatorily to imply something negative about the Muslim identity, differentiating and alienating Muslims in a subtle but damaging way. Moreover, we also provide an analysis of the toxicity of Islamophobic content; during the data collection campaign, we used semi-coded and OOV terms, but the content seems more toxic and racial compared to other forms of hate speech online, such as Antisemitism. 
\begin{figure*}[h]
    \centering
    \includegraphics[width=1\linewidth]{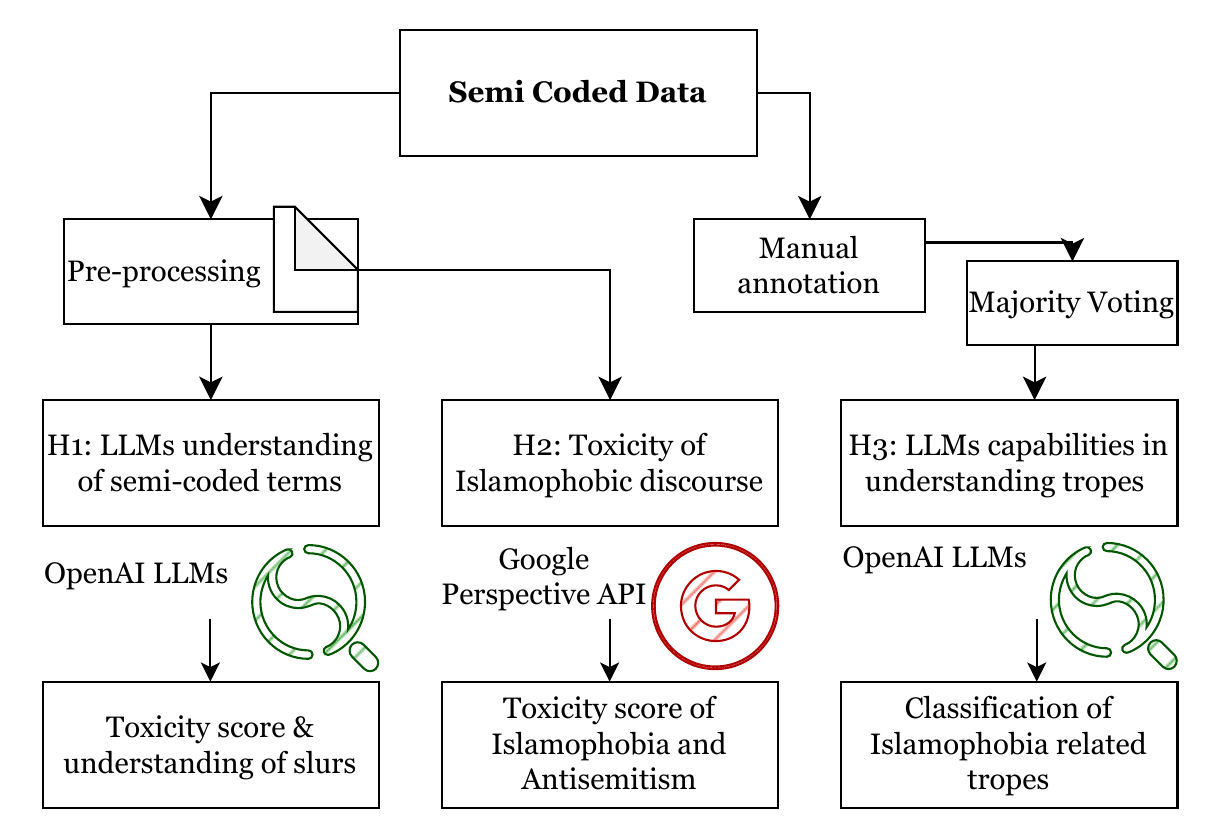}
    \caption{Pipeline for exploring Islamophobic narratives using topic modeling and LLM interpretation.}
    \label{fig:diagram}
     \vspace{-2mm}
\end{figure*}
\section{Dataset} 
\label{tab:dataset_labeling}
This work is an extension of a larger project entitled  \texttt{unmasking antisemitism} that our organization has undertaken to control anti-Jewish hatred using third-party software Pyrra\footnote{https://www.pyrratech.com/, however, now Pyrra is a part of Alertmedia - ( https://www.alertmedia.com/)}. Pyrra is a private
software company that allows its users to scrape posts from
alt-social media platforms according to a list of seed terms.
The dataset was collected between 2024 and 2025 and comprises nearly 63,628 instances. The antisemitic dataset is used from our prior work~\cite{mustafa2024monitoring}; and the complete detail about the seed-words (coded terms) is explained in the study. Whereas for Islamophobia, we initially started with well-known terms (muzrat, pislam, etc) to collect data for extremist social platforms, such as 4Chan, Gab, Telegram, etc. However, during manual labeling, annotators identified additional terms. Note: Often, Islamophobic and Antisemitic content appears together, such as (pislam, pisrael), where initial Islamophobic terms are identified during coded term discovery for antisemitism \cite{kikkisetti2024coded}. The complete dataset statistics are shown in Table \ref{tab:data-stats}, whereas the labeled datasets are shown in Table~\ref{tab:labeled-stats}. We have shown the top 10 platforms; however other platforms included are -- BoardReader, Minds, Rumble, Truth Social, 8kun, VKontakte, Disqus, GreatAwakening, Bluesky, Wimkin, Mastodon, Counter Social.

\begin{table}[ht]
\centering
\caption{Top 10 platforms by total posts, including a combined row for all remaining platforms.}
\begin{tabular}{|lrr|}
\hline
{Platform} & {Total} & {Percentage} \\
\hline
4chan & 54377 & 85.46\% \\
Gab & 2637 & 4.14\% \\
Incels & 1862 & 2.93\% \\
Bitchute & 735 & 1.16\% \\
Patriots & 719 & 1.13\% \\
GETTR & 594 & 0.93\% \\
Telegram & 592 & 0.93\% \\
Communities & 539 & 0.85\% \\
Odysee & 450 & 0.71\% \\
KiwiFarms & 269 & 0.42\% \\
\hline
\textbf{All Other Platforms} & \textbf{854} & \textbf{1.34\%} \\
\hline
\end{tabular}

\label{tab:data-stats}
\end{table}


To compare the toxicity of Islamophobic text, we use the baseline dataset of our previous work named \textit{unmasking antisemitism} \cite{mustafa2024monitoring}, which consists of 350 samples and contains all the known tropes floated on these platforms, such as (Jews are disloyal, Jews are greedy, Jews have too much power, Jews killed Jesus, etc.,). Similarly, we took 350 random samples from Islamophobia as shown in Figure \ref{fig:islam-jews-cmp}.

\begin{table}[ht]
\centering
\caption{Top 10 platforms by total posts (labeled), including a combined row for all remaining platforms.}
\begin{tabular}{|lrr|}
\hline
{Platform} & {Total} & {Percentage} \\
\hline
4chan & 1377 & 67.57\% \\
Gab & 223 & 10.94\% \\
GETTR & 188 & 9.22\% \\
Patriots & 41 & 2.01\% \\
Bitchute & 37 & 1.82\% \\
BoardReader & 28 & 1.37\% \\
Rumble & 25 & 1.23\% \\
Telegram & 20 & 0.98\% \\
Incels & 17 & 0.83\% \\
Minds & 15 & 0.74\% \\
\hline
\textbf{All Other Platforms} & \textbf{69} & \textbf{3.39\%} \\
\hline
\end{tabular}
\label{tab:labeled-stats}
\end{table}


Islamophobic content labeling is done through undergraduate researchers (political science) and experts with foundational knowledge of Islamic terminology, cultural contexts, and common stereotypes. Annotators were also provided with a coding statement to categorize the data into five tropes; as discussed below:
\subsection{Five Islamophobic Tropes}
\begin{itemize}
    \item Women's Oppression -- Perpetuating myths about Muslim women, such as describing Muslim women as \textit{femoids} trained like dogs, a dehumanizing comparison intended to strip them of agency and rights. Similarly, criticizing them for wearing burkas, contrasting it with Mary, the mother of Jesus.
    \item Prophet -- Profane remarks about Prophet Muhammad
    \item Islam -- Religious hate speech shifts focus to Islam as a belief system, often mocking its practices or associating it with violence or extremism.
    \item Immigrants -- Hateful discourse on Muslim immigration to Western countries.
    \item Racial Bias -- Hate speech often employs slurs and stereotypes to demean race, religion, or cultural practices. Attacks on race, for instance, may generalize all Muslims as belonging to specific racial groups such as Arabs, South Asians, or
Africans, using derogatory terms rooted in colonial or racial hierarchies.
\end{itemize}
\subsection{Islamophobia Coding Statement}
\begin{itemize}
    \item Does the comment portray Muslim women in a degrading or stereotypical way, such as mocking their attire, denying their autonomy, or dehumanizing them on the basis of gender and religion?
    \item Does the comment express disrespect toward Prophet Muhammad through insults, mockery, fabricated accusations, or other forms of blasphemous or inflammatory framing intended to provoke hostility?
    \item Does the comment attack Islam as a religion, framing its beliefs, practices, or teachings as inherently violent, irrational, inferior, or threatening in a way that goes beyond civil critique?
    \item Does the comment frame Muslim immigrants as threats, depicting them as dangerous, and/or culturally incompatible?
    \item Does the post use racialized stereotypes or slurs toward groups commonly associated with Islam (even if the association is incorrect, such as Sikhs)?
\end{itemize}
 During labeling, we also selected rows that used fewer toxic words, since offensive language and words are easily flaggable. Please note that at least two annotators independently labeled each item, and disagreements were adjudicated by a senior reviewer with expertise in hate-speech research to ensure consistency.\footnote{Upon acceptance of this paper, we will release the data and the labeled classes for each annotator.} However, Cohen's kappa score is \textit{0.8} with a strong agreement on labeled classes.

 For pre-processing, we use steps to normalize the text documents by converting uppercase to lowercase and replacing references. Next, we remove links from the text. However, we replace the links with the titles of the posts they lead to. Finally, we remove all special characters, white spaces, and line breaks using regular expressions.
The pipeline for exploring Islamophobic narratives is shown in Figure \ref{fig:diagram}. Once we have the dataset, we use a proportion of the dataset to demonstrate LLMs capabilities in understanding semi-coded slurs in \textbf{H1}. Next, we use the Google Perspective API to find the toxicity of Islamophobic discourse in \textbf{H2}; however, here we use all labeled samples. Finally, in \textbf{H3}, we use LLMs to classify tropes expressed in social posts. In the following sub-sections, we explain each hypothesis in detail. 
\begin{table*}

\centering
 \caption{GPT-4 response on classifying OOV terms used on extremist social media platforms.}
 
\begin{tabular}{|p{6.5cm}|p{1.5cm}|p{9cm}|}
\hline
Text & GPT score & Brief explanation by GPT            \\ \hline
He was born in a Muslim country. He grew up in a Muslim country. He's a {\color{red}{muzrat}} \vspace{0.1cm} & 0.9               & \parbox{9cm}{\vspace{0.1cm} The text contains a high level of hate speech and negative sentiment towards Muslims.  The term {\color{red}{muzrat}} are derogatory term used against Muslims,  showing a clear intention to offend and spread hate. \vspace{0.1cm}}                                                                                                                                                                                                                                                                                              \\ \hline
\parbox{6.5cm}{\vspace{0.1cm}Having contributed more to humanity than modern Italy The Bernoullis alone did more for humanity than every {\color{red}{mudslime}} since the golden age ended\vspace{0.1cm}} & 0.85              & \parbox{9cm}{\vspace{0.1cm}The text contains derogatory slang terms {\color{red}{mudslime}}, which are used as hate speech towards Muslims. The score is high because of the extreme hateful language used.\vspace{0.1cm}}                                                                                                                                                                                                                                                                                                                                   \\ \hline
\parbox{6.5cm}{Sure {\color{red}{abdul}}, but I read her name so I know she's your {\color{red}{mudslime}} kind }                                                                                                                       & 1                 & \parbox{9cm}{\vspace{0.1cm}This text contains explicit hate speech towards Muslims. The term {\color{red}{mudslime}} is a derogatory term used to insult Muslims  and is indicative of extreme negativity and hate.\vspace{0.1cm}}                                                                                                                                                                                                                                                                                                                          \\ \hline
Is he wrong though? Sorry Abdul, there’s no place for {\color{red}{muzzies}} in Trump’s world                                                                                                            & 1                 & \parbox{9cm}{\vspace{0.1cm}This text contains extreme hate speech towards Muslims. The term {\color{red}{muzzies}} is used as a derogatory term for Muslims,  and the statements imply that Muslims are not welcome and are a threat.\vspace{0.1cm}}                                                                                                                                                                                                                                                                                                        \\ \hline
I do not own anything {\color{red}{pislam}} in my home, and I would burn it if I did.                                                                                                                    & 1                 & \parbox{9cm}{\vspace{0.1cm}The text shows extreme hate speech towards Muslims.  The author uses the derogatory term {\color{red}{pislam}} to insult Islam and suggests burning any related objects in their home.\vspace{0.1cm}}                                                                                                                                                                                                                                                                                                                           \\ \hline
Their {\color{red}{Mohammedan}} friends will give them a quick boost at the start.                                                                                                                       & 0.1               & \parbox{9cm}{\vspace{0.1cm}This text does not contain hate speech towards Muslims or any form of negativity.  The term {\color{red}{Mohammedan}} is outdated and some might consider it disrespectful  because it incorrectly implies that Muslims worship Muhammad,  which they do not, they worship God (Allah). However, it doesn't seem to be used with harmful intent in this context. The overall message of the text seems to be neutral or even slightly positive,  as it talks about these friends giving "a quick boost.""\vspace{0.1cm}}  \\ \hline
France will probably be the first country to fall to the {\color{red}{mohammedan}} wretches.                                                                                                             & 0.9               & \parbox{9cm}{\vspace{0.1cm}This text is a clear example of hate speech.  It uses derogatory language (``wretches'') to dehumanize and demean Muslims ({\color{red}{Mohammedan}}), and it assumes and promotes a negative outcome (France ``falling'') due to the presence of Muslims.\vspace{0.1cm}}                                                 \\ \hline
\end{tabular}
\label{tab:gpt-4-response}
\end{table*}

\section{Results and Discussion} \label{tab:results}
\subsection{H1: LLMs Capabilities of OOV and Semi-coded Terms in Islamophobia}\label{tab:h1}
Islamophobic discourse shifted toward OOV (semi-coded) terms like \textit{mohammedan} and \textit{mudslime}, among others, that imply threats without explicitly violating platform guidelines. Such language often appears in memes, jokes, or seemingly neutral phrases that subtly convey harmful stereotypes. The main types of hate speech involve lexical manipulation and contextual misuse of words. Lexical manipulation here would be muzrat, as explained, this is the changing of the prefix and suffix of a word to combine a neutral word and an offensive word to create a new hate term. Contextual misuse would be the example of Mohammedan, which is a neutral word in general, that becomes hateful when the user manipulates the context to convey a negative connotation. We also argue that recent advancements in LLMs made extracting slurs easier \cite{kikkisetti2024coded} and that LLMs such as GPT-4 understand the toxicity of such messages. Therefore, we selected social posts where terms \textit{mudslime, mohammedan, pislam, muzzies and muzrats} have been used and posts contain less toxicity. Most data include such terms with abusive language, which we believe LLMs can easily flag. The prompt we use is as follows:

\noindent\textbf{Prompt: }\textit{Analyze the following text and assign a score between 0 and 1, where 0 represents no hate or negativity, and 1 represents extreme hate speech toward Muslims. Provide a score along with a brief explanation.}

We present the GPT-4 negativity score, along with a brief explanation, in Table \ref{tab:gpt-4-response}. It is clear from these responses and negativity scores that advanced LLMs understand these terms. However, the term \textit{mohammedan} achieves reasonable accuracy, whereas GPT-4 relies more on context to categorize negativity. For instance, in 6\textsuperscript{th} post, LLMs response states: \textit{The term mohammedan is outdated and some might consider it disrespectful}, whereas in 7\textsuperscript{th} post (comment), where the word \textit{wretches}, changes the negativity score of the sentence and considers \textit{mohammedan} as a negative term used against Muslims. Additionally, \textit{abdul} is ignored as a hate term in the third example; the LLM likely assumed this to be a reference to the target's name, which the human annotators know is not the case. The term \textit{abdul} is in the same position as \textit{mohammedan} as both are just named that rely on context to make it offensive to use. This suggests that if \textit{mudslime} was removed from the sentence, GPT would have a harder time labeling this correctly as hate speech, and the score would have been lower. However, from these responses, it is clear that LLMs understand the terms and measures required by social platforms to prevent the spread of such terms. 

To evaluate whether GPT-4 consistently assigns a high hate score to Islamophobic slurs, we compared the mean score against a neutral baseline of 0.5. The mean score $M$ is ($0.8767$). A one-sample $t-test$ showed that this mean was significantly higher than the baseline ($t = 10.78$, $p = 3.15 \times 10^{-14}$). Since the p-value is below $\alpha =  0.05$, indicating that GPT-4 strongly and consistently identifies these slurs as highly negative and hateful -- see Table \ref{tab:h1-stats}


\begin{table}[h!]
\centering
\caption{Statistical validation of GPT-4 scores for Islamophobic slurs.}
\begin{tabular}{|lll|}
\hline
{Statistic} & {Value} & {Interpretation} \\
\hline
Mean Toxicity & 0.8767 & High average toxicity \\
t-statistic & 10.7786 & Significant difference -- baseline \\
$p$-value (t-test) & $3.15 \times 10^{-14}$ & Strong evidence against null \\
\hline
\end{tabular}
\label{tab:h1-stats}
\end{table}

\begin{table*}
\centering
\caption{Toxicity samples from the highest frequency words of Antisemitism and Islamophobia.}
\begin{tabular}{|p{15cm}|p{2.5cm}|}
\hline
\multicolumn{1}{|l|}{Text}                                                                                                                                                                                                                                                                                                & Toxicity Score \\ \hline
\multicolumn{2}{|l|}{Antisemitism}                                                                                                                                                                                                                                                                                                         \\ \hline
\parbox{15cm}{ \vspace{0.1cm} the liberal world order also known a globalists the cabal and the zionist ha declared full war on america \\ and on any christian conservative nation hungary poland russia brazil or we fight and defeat them once \\ and for all or we become their slave \vspace{0.1cm}} & 0.37           \\ \hline
\parbox{15cm}{ \vspace{0.1cm}why would our zionist occupied government arrest one of their own \vspace{0.1cm}}                                                                                                                                                                                                                                   & 0.30           \\ \hline
\multicolumn{2}{|l|}{Islamophobia}                                                                                                                                                                                                                                                                                                         \\ \hline
\parbox{15cm}{ \vspace{0.1cm}Every time a church burns... Burn every mosque in the country and deport every raghead in the country \vspace{0.1cm}}                                                                                                                                                                                               & 0.65           \\ \hline
\parbox{15cm}{ \vspace{0.1cm}All the shitskins playing shitskin olympics, I hate all of you equally and want you all out of my country \vspace{0.1cm}}                                                                                                                                                                                           & 0.90           \\ \hline
\end{tabular}
\label{tab:h2-examples}
\end{table*}
\begin{figure}[h]
    \centering
    \includegraphics[width=1\linewidth]{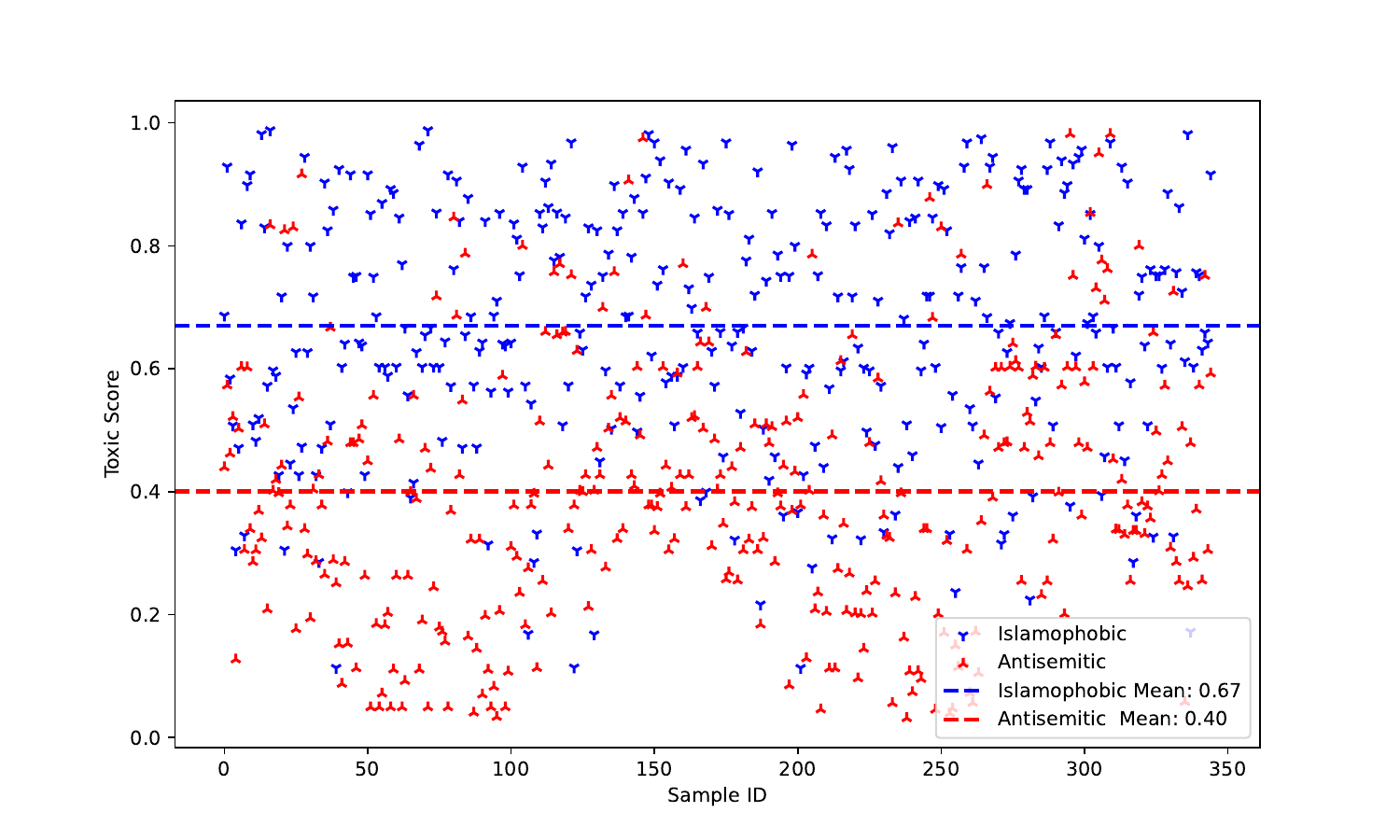}
    \caption{Toxicity comparison of Islamophobia and Antisemitism.}
    \label{fig:islam-jews-cmp}
\end{figure}

\begin{figure*}[t]
    \centering

    \subfigure[Racism]{%
        \includegraphics[width=0.32\textwidth]{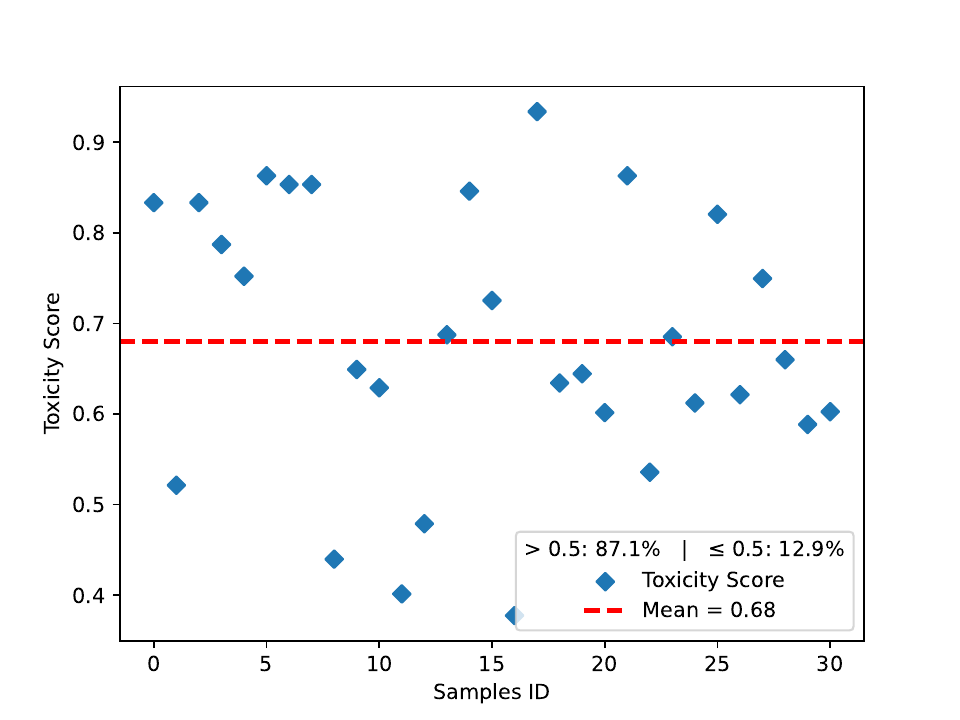}
    }\hfill
    \subfigure[Immigrants]{%
        \includegraphics[width=0.32\textwidth]{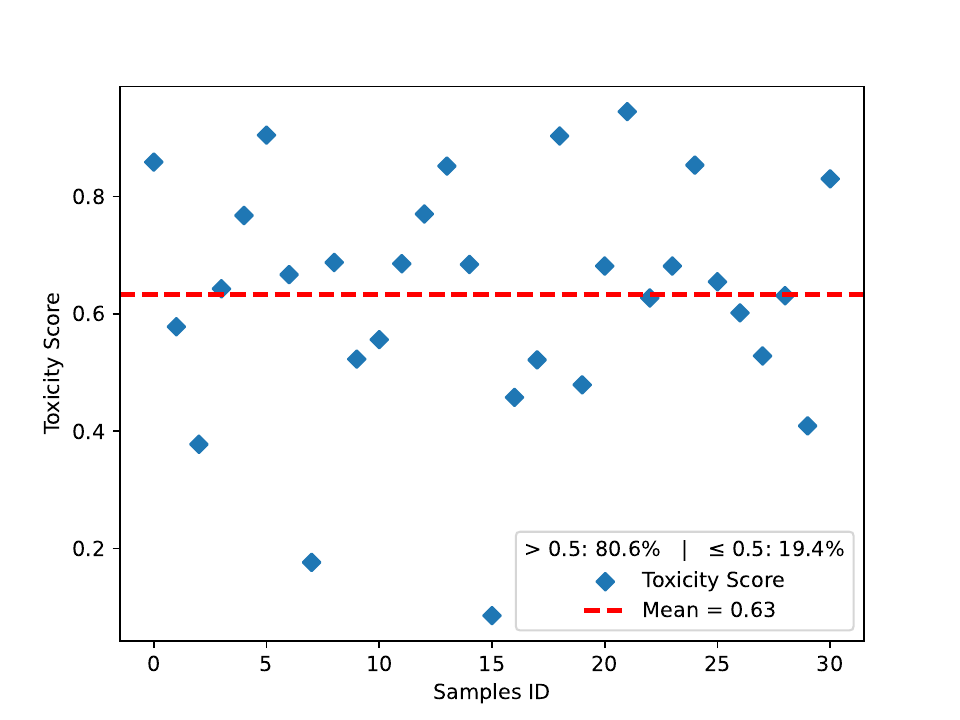}
    }\hfill
    \subfigure[Islam]{%
        \includegraphics[width=0.32\textwidth]{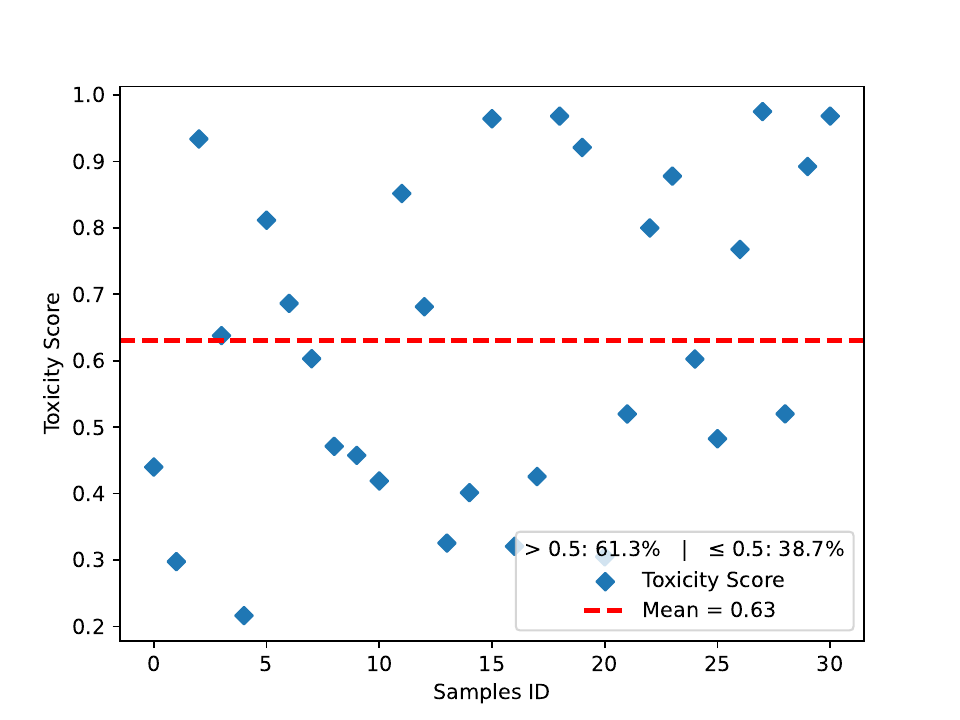}
    }

    \subfigure[Muhammad]{%
        \includegraphics[width=0.32\textwidth]{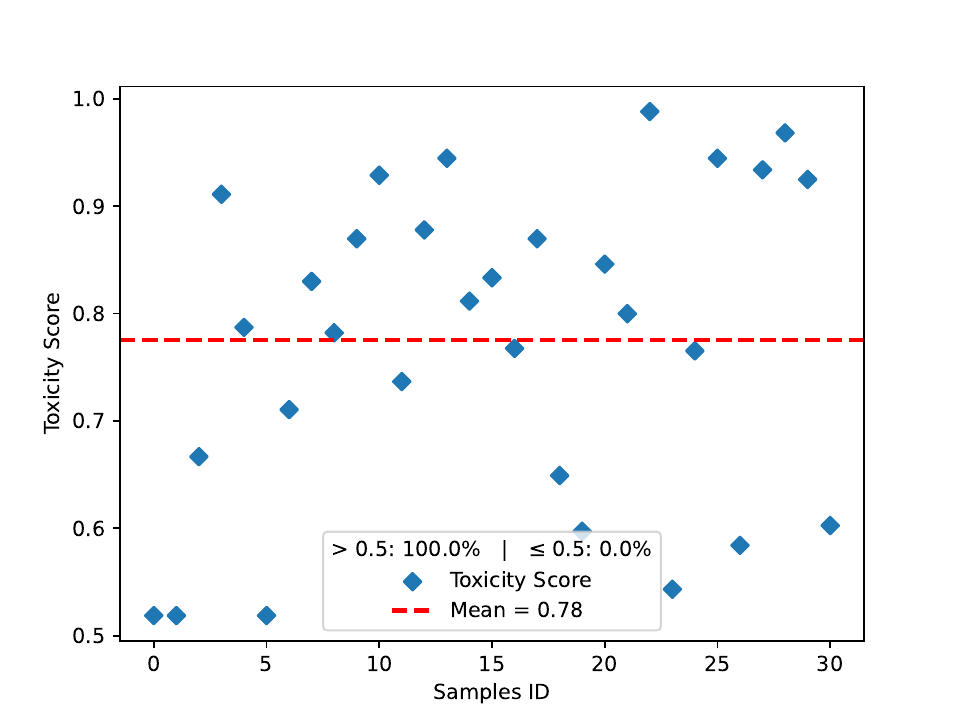}
    }
        \subfigure[Oppression]{%
        \includegraphics[width=0.32\textwidth]{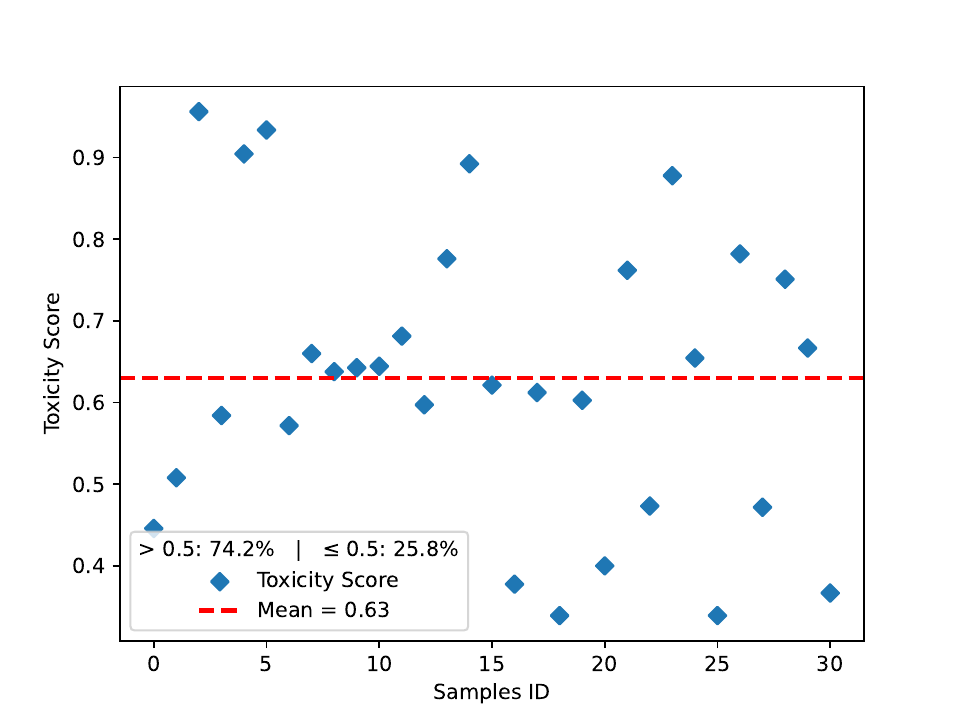}
    }
    \caption{Toxicity comparison of different tropes in Islamophobia.}
    \label{fig:toxic_tropes_islam}
\end{figure*}

\subsection{H2: Hate against Muslim are more Toxic} \label{tab:toxic}
In this section, we compare the toxicity of Islamophobia and Antisemitism. We use Google Perspective API \cite{GooglePerspectiveAPI} to annotate each message in our dataset with
a score that reflects how rude or disrespectful a comment is.  The Google Perspective API is a widely adopted ML-based toxicity-scoring tool that estimates the \textit{perceived impact} of a given text on conversational civility. Previous work has also validated the performance
of the Perspective API~\cite{gehman2020realtoxicityprompts}.

We use baseline data on Antisemitism to compare it to Islamophobia. We chose to analyze Antisemitism because Jewish people are also being targeted for hate crimes due to their religious, ethnic, and cultural identity.  We randomly selected samples from both kinds of hate speech and considered the most recent tropes floated on social media, such as Islamization and Jews controlling the world economy. We showed the toxicity score in Figure \ref{fig:islam-jews-cmp}, where the Y-axis shows the toxicity score and the X-axis represents our samples. From figure~\ref{fig:islam-jews-cmp}, the mean value toxicity of all the Islamophobic samples is 0.67, whereas the mean value of Antisemitism is 0.40. In addition to comparing the overall toxicity of Islamophobic posts, we show class-wise comparison as well, see Figure \ref{fig:toxic_tropes_islam} a) as Racism, b) Immigrants, c) Islam (hate against religion; where discussion is purely about religion), d) Hate against Prophet Muhammad, and e) Oppression of women in Islam. We show the toxicity score on the Y-axis and 30 samples of each class on the X-axis. We also provide the mean value and the percentage of samples above and below the toxicity standardized threshold of 0.5.
%
We observe hate against Immigrants and Prophet Muhammad is more toxic compared to other tropes in Islamophobia as can be seen in Figures \ref{fig:toxic_tropes_islam} (b) and (c).

\begin{figure*}[h]
    \centering
    \includegraphics[width=1\linewidth]{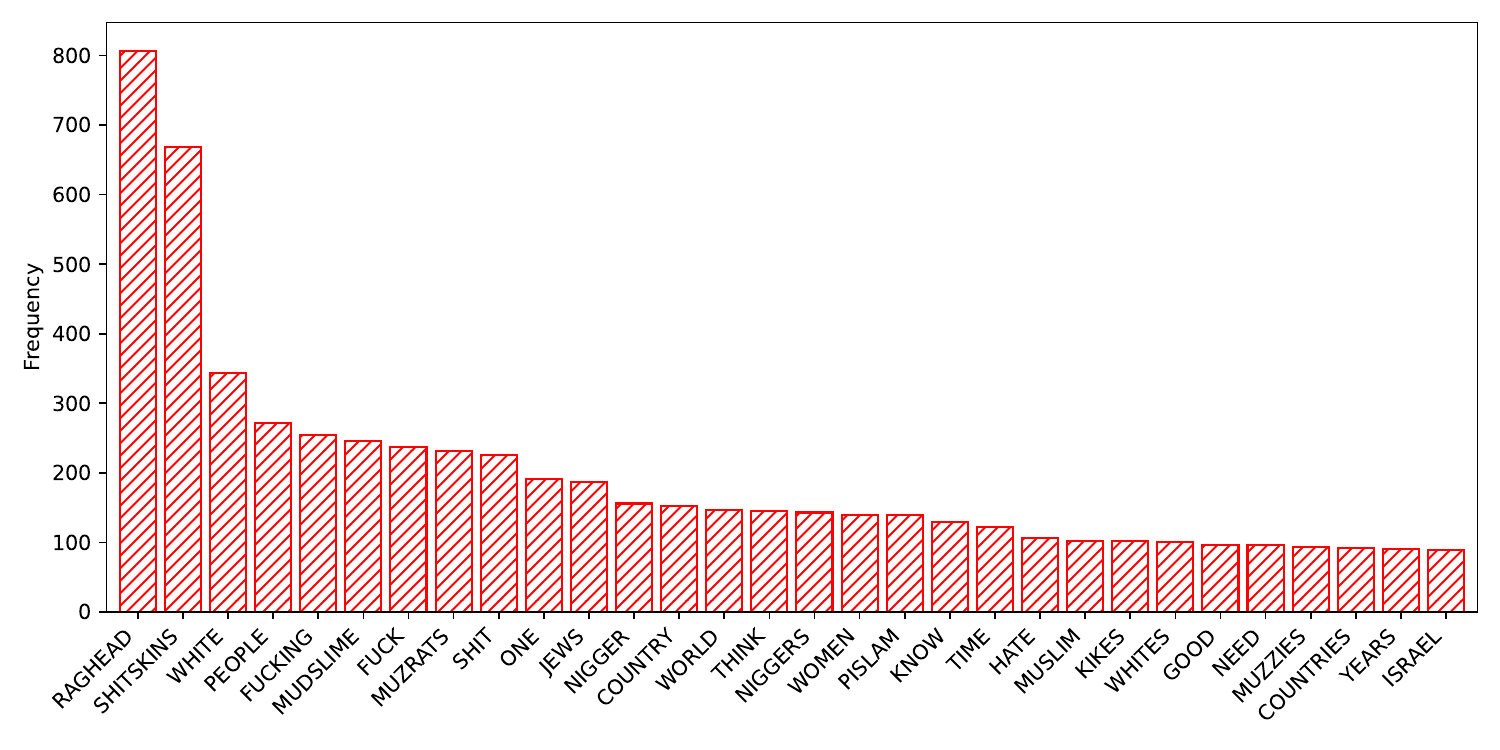}
    \caption{30 most occurring words in Islamophobia.}
    \label{fig:islam-tag-cloud}
\end{figure*}

\begin{figure*}[h]
    \centering
    \includegraphics[width=1\linewidth]{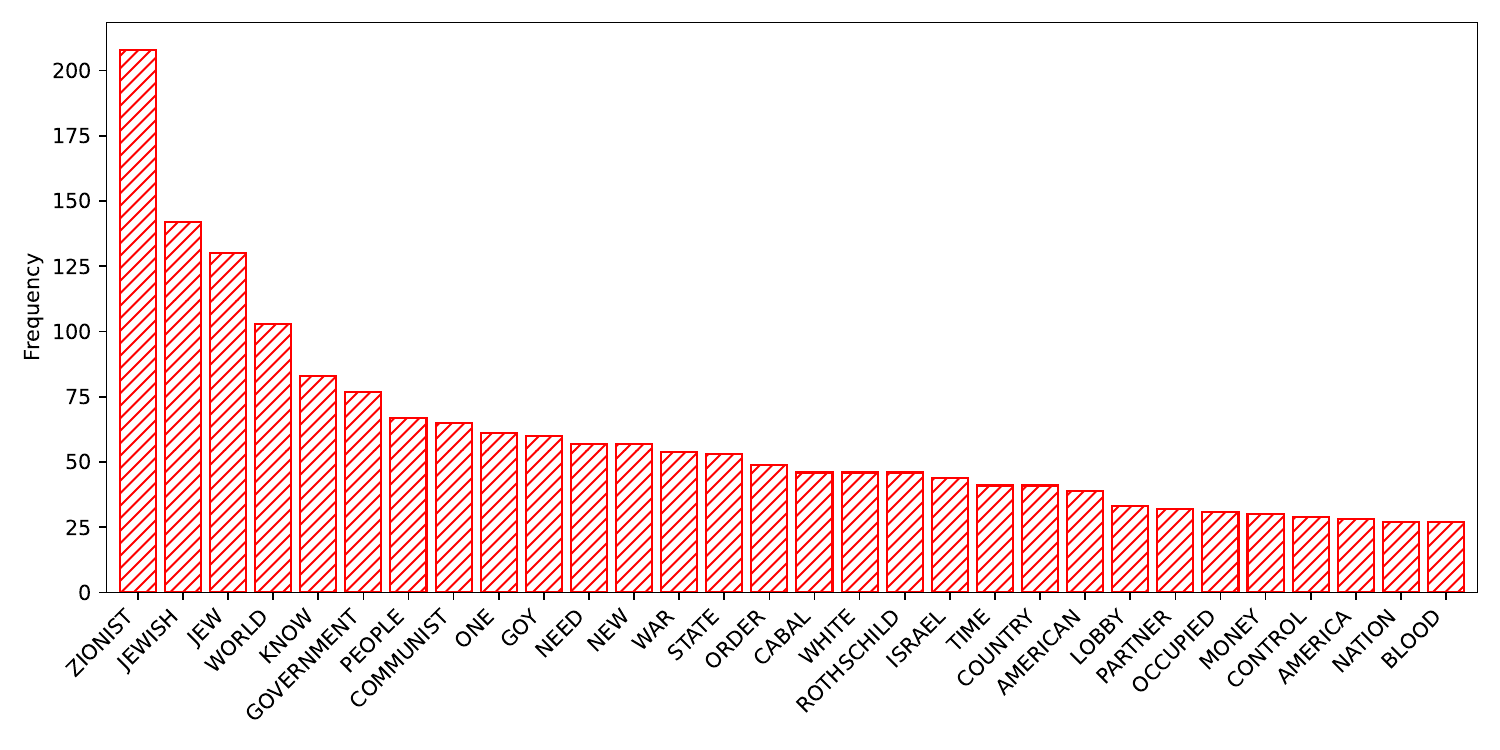}
    \caption{30 most occurring words in Antisemitism.}
    \label{fig:jews-tag-cloud}
\end{figure*}

We also show top 30 words in both Islamophobia and Antisemitism in Figure \ref{fig:islam-tag-cloud} and \ref{fig:jews-tag-cloud} and a few samples in Table \ref{tab:h2-examples}. 
The most floated term in Islamophobia is \textit{raghead, shitskins and muzrats}, whereas in Antisemitism the terms are \textit{jewish, zionist, world}. 
The term \textit{raghead} is a derogatory slur used to mock Muslims for their traditional head coverings. While Muslim women frequently wear the hijab, Muslim men across different cultures wear a variety of head coverings such as keffiyehs, kufis, and taqiyahs—many of which are rooted in cultural rather than strictly religious traditions. This slur falsely implies that covering one’s head is inherently tied to extremism, reinforcing the stereotype that Muslims, particularly those who visibly adhere to Islamic customs, are terrorists. Whereas the term \textit{sh*tskin} is a deeply xenophobic and racist slur historically used against Muslims. This term attacks the perceived racial identity of Muslims, reflecting the stereotype that Muslims are predominantly Arab, South Asian, or African. By emphasizing skin color as a marker of inferiority, this slur exposes the racial dimension of Islamophobia, illustrating how anti-Muslim hate often intersects with broader racist ideologies. We show the use of the term \textit{Muzrat} in section \ref{tab:dterms}.

Similarly, in Antisemitism, the term \textit{world} is associated with \textit{New World Order}. It is an Antisemitic term when it is followed by a reference to a Jewish business leader or political official with a secret agenda to take control of the world. Finally, Zionism is a movement and ideology to reestablish and support the existence of a Jewish state in the Biblical Land of Israel \cite{kikkisetti2024coded}. 
To statistically validate our hypothesis, we conducted a two-sample t-test comparing the mean toxicity scores of content targeting Muslims and Jews. The results show that the mean toxicity associated with Muslim-targeted hate is significantly higher than that of antisemitic hate (t = 17.01, $p \approx 0$), providing strong evidence that Islamophobic content exhibits a higher mean toxicity score than antisemitic content. These results are demonstrated in Table~\ref{tab:stat-h2}.

%

\begin{table}[h!]
\centering
\caption{Statistical Test Results}
\begin{tabular}{|lcc|}
\hline
{Statistic} & {Value} & {p-value} \\
\hline
t-statistic & 17.01 & $9.76 \times 10^{-55}$ \\
\hline
\end{tabular}
\label{tab:stat-h2}
\end{table}

\begin{table*}[]
\centering
\caption{Tropes examples}
\begin{tabular}{|l|l|}
\hline
Text& Trope               \\ \hline
\parbox{12cm}{\vspace{0.1cm}neither one of those freaks are native american  indians . camelhair is   communist india and and raghead.}
 & Racism              \\ \hline
we are clearly superior to shitskins like you, though that isn't a particularly high bar.& Racism              \\ \hline
\parbox{12cm}{\vspace{0.1cm}frankly i don't want people to get sidetracked. get rid of all the shitskins first and foremost } & Immigration         \\ \hline
\parbox{12cm}{\vspace{0.1cm}muzrats are stealing your whole country.} & Immigration         \\ \hline
\parbox{12cm}{\vspace{0.1cm}the persians were high iq. pisslam moved in and it became an inbred shithole. }& Islam            \\ \hline
\parbox{12cm}{\vspace{0.1cm}you need to be the product of at least 10 generations of inbreeding to be susceptible to pislam. }& Islam            \\ \hline
\parbox{12cm}{\vspace{0.1cm}the prophet muhammad was a pedophile and a rapist. the muzzies worship him}& Muhammad             \\ \hline
\parbox{12cm}{\vspace{0.1cm}are you that mudslime that lives in the Netherlands \\ what do you think about pedophile prophet (piss be upon him) amen}                            & Muhammad             \\ \hline
\parbox{12cm}{\vspace{0.1cm}we don't live in the 20th century anymore ahmed, it's very rare and illegal for priests to f*** churchboys nowadays whereas it's perfectly normal and accepted to wife nine year old girls in raghead countries.}                                 & Oppression\\ \hline
\parbox{12cm}{\vspace{0.1cm}all women need wear a burka thats promotion of pissla maryan women do not cover their faces or hair,  we are proud of the beauty of our race throughout all european history} & Oppression\\ \hline
\end{tabular}
\label{tab:tropes-examples}
\end{table*}
\begin{table*}[h!]
\centering
\caption{Islamic Tropes Classification by OpenAI – GPT-4.}
\begin{tabular}{|l|l|l|l|}
\hline
Trope                                & Precision & Recall & F-score \\ \hline

Racism                & 0.83   &   0.60&      0.70      \\ \hline
Muslim immigrants & 0.59   &   0.48      & 0.53       \\ \hline
Islam        & 0.67  &    0.88&      0.76    \\ \hline
Prophet Muhammad                & 0.86  &    0.90  &    0.88       \\ \hline
Perpetuating stereotypes about the oppression of women in Islam                & 1.00     & 0.43    &  0.60       \\ \hline

\end{tabular}
\label{tab:open-ai-islam}
\end{table*}
\begin{figure}
    \centering
    \includegraphics[width=1\linewidth]{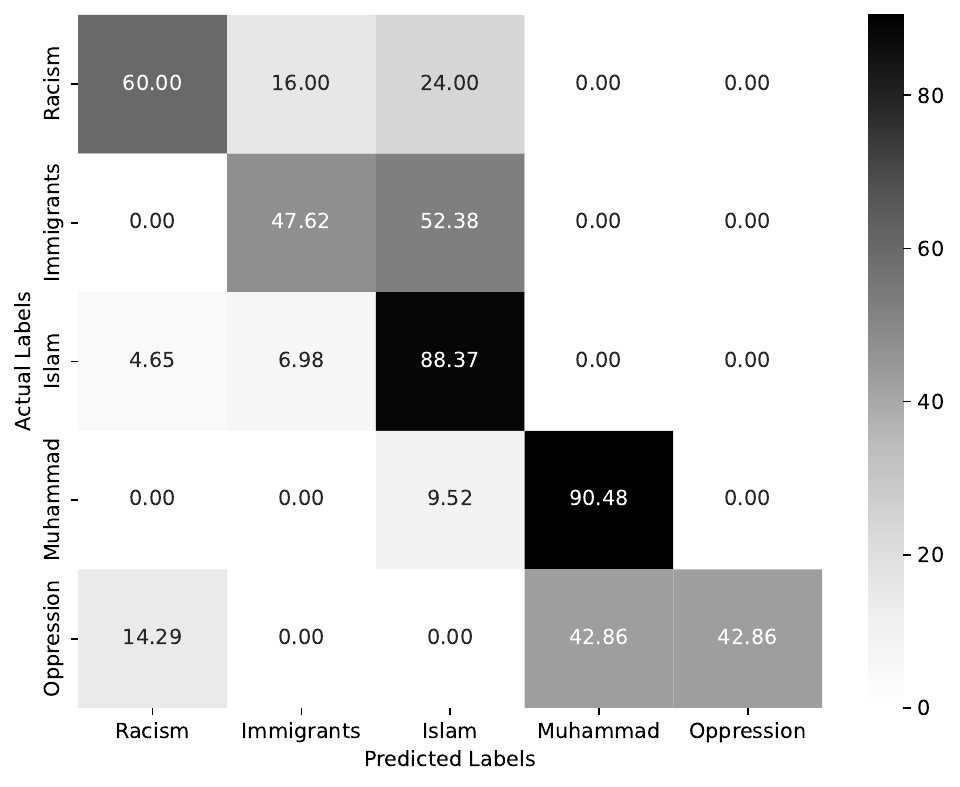}
    \caption{Confusion matrix -- OpenAI GPT on Islamic tropes. }
    \label{fig:cmislam}
\end{figure}
\subsection{H3: Islamophobic tropes classification} \label{tab:tropes-islam-h3}


In this section, we provide results of GPT-4 classification results. We also provide two examples of each trope shown in Table \ref{tab:tropes-examples}. 
\subsubsection{Racism}
The posts in Table \ref{tab:tropes-examples} uses several racial/ethnic slurs (\textit{camelhair}, \textit{raghead} that are commonly directed toward people of Middle Eastern or South Asian descent. Although the statement does not explicitly reference Islam, these racialized slurs are widely used for Muslims, revealing how Islamophobia is often entangled with racism and ethnonationalism. The comment also asserts that certain individuals are not \textit{Native to America,} implying a gatekeeping of national identity and positioning racialized others as illegitimate occupants of the country.
Therefore, the dominant trope here is Racial Bias.
\subsubsection{Immigration}
The term muzrat is combined with the claim that Muslims are \textit{stealing your whole country} invokes demographic replacement and invasion narratives, suggesting that Muslim presence is an existential threat to national identity. This example clearly fits the Immigration trope, framing Muslims as foreign infiltrators. 
\subsubsection{Islam}
The comments constructs a contrast between \textit{high IQ Persians} and a degraded society supposedly resulting from the arrival of \textit{pisslam,} a derogatory slur for Islam. The speaker attributes social decline, \textit{inbreeding,} and reduced intelligence to the influence of Islam as a religion, not a specific individual or ethnic group.
Importantly, the argument frames Islam as a corruptive force that inherently damages civilizations. The reference to \textit{inbreeding} also aligns with long-standing stereotypes about Islamic marriage practices (e.g., cousin marriage), used here to signal degeneracy. Thus, this fits the Islam (Religion) trope because the attack is on Islamic belief and its alleged societal effects.
\subsubsection{Muhammad}
The posts in this category as shown in Table \ref{tab:tropes-examples} directly insults Prophet Muhammad by assigning defamatory labels (\textit{pedophile,} \textit{rapist}). These accusations draw from well-known Islamophobic narratives that distort historical accounts to justify hostility toward Muslims. Because Prophet Muhammad is central to Islamic identity, attacking his moral character functions as a direct form of religious defamation.
The phrase \textit{the muzzies worship him} further reinforces hostility by pairing a slur with an implied irrational devotion. Hence, this is classified under the Prophet trope.

\subsubsection{Oppression}
In the first example shown in Table \ref{tab:tropes-examples}, the comment employs an oppression-of-women trope by invoking the \textit{nine-year-old wife} stereotype, which falsely portrays Muslim women as subject to normalized child marriage. Next sentence targets Muslim women by mocking the practice of wearing the burka. The use of \textit{pissla} (pislam) signals that the speaker views Islamic modesty practices as inherently negative. 
%
Thus, this goes under the oppression trope. \\

We use the following prompt to categorize the posts into tropes.
\\
\noindent\textbf{Prompt: }\textit{The provided texts contain coded and out-of-vocabulary language that expresses Islamophobic sentiments, spreading hate against Muslims on social media. Classify the text sentiments into one of the following categories: i) Racial prejudice towards Muslims, ii) Hate Speech Against Muslim Immigrants, iii) Hate Speech Against Muslim Religion, iv)  Hate Speech Against Muslim Prophet Muhammad, and v) Hate Speech Perpetuating Stereotypes about the Oppression of Women in Islam}

We then use OpenAI GPT-4 predictions against a human-labeled dataset to compute a confusion matrix and a classification report. This enables us to comprehend the capabilities of advanced AI in trope discovery and understanding of Islamophobia. The results of the confusion matrix and classification report, such as Precision, Recall, and F-score, are shown in Figure \ref{fig:cmislam} and Table \ref{tab:open-ai-islam}, respectively. Precision measures the number of flagged posts that are hateful, recall measures the number of hateful posts that were correctly flagged, and F-score balances both to reflect overall detection accuracy \cite{mustafa2017early}. From the confusion matrix, it is clear that advanced LLMs struggle to understand tropes involving Oppression, Racism, and Immigration, whereas the other two tropes showed better results. While investigating, the reason for the better result is the frequent use of the words -- \textit{pedophile, mohammed pedophile} for the trope (Muhammad); whereas \textit{mudslime} is found in (Islam). As per our observation, we find that the OOV term \textit{mudslime} is associated with the trope where hate speech is targeted explicitly at Islam as a religion. Similarly, OOV term \textit{pislam} is frequently found in both trope \textit{Prophet and Islam} as it can also be seen as an alternative use (piss be upon him) in Table \ref{tab:tropes-examples}. Our findings on AI capabilities are as follows:
\begin{itemize}
    \item Although advanced LLMs understand most of the OOV and Neutral terms, they show limitations in detecting ideologically driven content.
    \item Use of certain words and bi-grams also differentiates a few tropes, for instance, the frequent use of the words \textit{pedophile, piss be upon him} for trope (Prophet).
\end{itemize}

\section{Conclusion and Future Work} \label{tab:conclusion}
In this work, we performed one of the first analysis of Out of Vocabulary (OOV) and semi-coded terms such as (muzrat, pislam, mudslime, mohammedan, muzzies), which are used to spread hate against Muslims using disrespectful and unreasonable posts. Using LLMs like GPT-4, we demonstrated their capabilities to understand these terms and tropes. Our findings indicate that LLMs recognize these words as being disrespectful to Muslims; however, they struggle to distinguish the theme expressed. We find that the toxicity of Islamophobic hate speech is much higher than that of other kinds of hate speech, like Antisemitism. The choice of Antisemitism is based on the fact that often, both appear to show rude discourse on these extremist social platforms by far-right activists.
Our results have several implications for the research community working in the hate speech domain: i) First, such results can assist social media platforms, law enforcement agencies, and policymakers in better understanding and mitigating online toxicity of Islamophobia ii) Real-world hate speech variations across different extremist social platforms, iii) Balancing freedom of speech vs. hate speech regulation. Future work includes examining other types of hate slurs, such as \textit{pajeet, chink, poos}, identified during this study, and training an AI model to detect such terminologies for a safer online community.

\ifCLASSOPTIONcaptionsoff
  \newpage
\fi

\bibliographystyle{ieeetr}
\textbf{\bibliography{main}}

@article{awan2016islamophobia,
  title={Islamophobia on social media: A qualitative analysis of the facebook's walls of hate},
  author={Awan, Imran},
  journal={International Journal of Cyber Criminology},
  volume={10},
  number={1},
  pages={1},
  year={2016},
  publisher={International Journal of Cyber Criminology}
}

@article{gehman2020realtoxicityprompts,
  title={Realtoxicityprompts: Evaluating neural toxic degeneration in language models},
  author={Gehman, Samuel and Gururangan, Suchin and Sap, Maarten and Choi, Yejin and Smith, Noah A},
  journal={arXiv preprint arXiv:2009.11462},
  year={2020}
}

@online{meta_hateful_conduct_policy,
  title        = {Meta Transparency Center – Policy: Hateful Conduct (Community Standards)},
  author       = {{Meta Platforms, Inc.}},
  year         = {2025},
  month        = {1},
  url          = {https://transparency.meta.com/policies/community-standards/hateful-conduct/},
  organization = {Meta Transparency Center},
  urldate      = {2025-08-17}
}

@online{x_hateful_conduct_policy,
  title        = {X's Policy on Hateful Conduct},
  author       = {{X Corp.}},
  year         = {2023},
  month        = {4},
  url          = {https://help.x.com/en/rules-and-policies/hateful-conduct-policy},
  organization = {X Help Center},
  urldate      = {2025-08-17}
}

@inproceedings{davidson2017automated,
  title={Automated hate speech detection and the problem of offensive language},
  author={Davidson, Thomas and Warmsley, Dana and Macy, Michael and Weber, Ingmar},
  booktitle={Proceedings of the international AAAI conference on web and social media},
  volume={11},
  number={1},
  pages={512--515},
  year={2017}
}

@inproceedings{hine2017kek,
  title={Kek, cucks, and god emperor trump: A measurement study of 4chan’s politically incorrect forum and its effects on the web},
  author={Hine, Gabriel and Onaolapo, Jeremiah and De Cristofaro, Emiliano and Kourtellis, Nicolas and Leontiadis, Ilias and Samaras, Riginos and Stringhini, Gianluca and Blackburn, Jeremy},
  booktitle={Proceedings of the International AAAI Conference on Web and Social Media},
  volume={11},
  number={1},
  pages={92--101},
  year={2017}
}

@article{boyuk2024hate,
  title={Hate Speech on Social Media in the Axis of Islamophobia: Example of TikTok},
  author={B{\"o}y{\"u}k, Mustafa},
  journal={Journal of Media and Religion Studies},
  number={1st International Media, Digital Culture and Religion Congress Special Issue},
  pages={91--122},
  year={2024},
  publisher={Hakan AYDIN}
}

@inproceedings{wasi2024explainable,
  title={Explainable identification of hate speech towards islam using graph neural networks},
  author={Wasi, Azmine Toushik},
  booktitle={Proceedings of the Third Workshop on NLP for Positive Impact},
  pages={250--257},
  year={2024}
}

@article{rosen2025antisemitic,
  title={Antisemitic and Islamophobic hate speech precedes a decrease in lexico-semantic diversity in comment threads online},
  author={Rosen, ZP and Dale, Rick},
  journal={Humanities and Social Sciences Communications},
  volume={12},
  number={1},
  pages={1--14},
  year={2025},
  publisher={Palgrave}
}

@article{sen2024hatetinyllm,
  title={Hatetinyllm: Hate speech detection using tiny large language models},
  author={Sen, Tanmay and Das, Ansuman and Sen, Mrinmay},
  journal={arXiv preprint arXiv:2405.01577},
  year={2024}
}

@inproceedings{ghorbanpour2025can,
  title={Can prompting llms unlock hate speech detection across languages? a zero-shot and few-shot study},
  author={Ghorbanpour, Faeze and Dementieva, Daryna and Fraser, Alexandar},
  booktitle={Proceedings of the The 9th Workshop on Online Abuse and Harms (WOAH)},
  pages={413--425},
  year={2025}
}

@inproceedings{roy2023probing,
  title={Probing LLMs for hate speech detection: strengths and vulnerabilities},
  author={Roy, Sarthak and Harshvardhan, Ashish and Mukherjee, Animesh and Saha, Punyajoy},
  booktitle={Findings of the association for computational linguistics: EMNLP 2023},
  pages={6116--6128},
  year={2023}
}

@article{mustafa2017early,
  title={Early detection of controversial urdu speeches from social media},
  author={Mustafa, Raza Ul and Nawaz, M Saqib and Farzund, J and Lali, Muhammad Ikram and Shahzad, Basit and Viger, PF},
  journal={Data Sci. Pattern Recognit},
  volume={1},
  number={2},
  pages={26--42},
  year={2017}
}

@article{burnap2015cyber,
  title={Cyber hate speech on twitter: An application of machine classification and statistical modeling for policy and decision making},
  author={Burnap, Pete and Williams, Matthew L},
  journal={Policy \& internet},
  volume={7},
  number={2},
  pages={223--242},
  year={2015},
  publisher={Wiley Online Library}
}

@article{watanabe2018hate,
  title={Hate speech on twitter: A pragmatic approach to collect hateful and offensive expressions and perform hate speech detection},
  author={Watanabe, Hajime and Bouazizi, Mondher and Ohtsuki, Tomoaki},
  journal={IEEE access},
  volume={6},
  pages={13825--13835},
  year={2018},
  publisher={IEEE}
}

@inproceedings{zsisku2024hate,
  title={Hate speech detection and reclaimed language: Mitigating false positives and compounded discrimination},
  author={Zsisku, Eszter and Zubiaga, Arkaitz and Dubossarsky, Haim},
  booktitle={Proceedings of the 16th ACM Web Science Conference},
  pages={241--249},
  year={2024}
}

@inproceedings{qu2023unsafe,
  title={Unsafe diffusion: On the generation of unsafe images and hateful memes from text-to-image models},
  author={Qu, Yiting and Shen, Xinyue and He, Xinlei and Backes, Michael and Zannettou, Savvas and Zhang, Yang},
  booktitle={Proceedings of the 2023 ACM SIGSAC conference on computer and communications security},
  pages={3403--3417},
  year={2023}
}

@misc{GooglePerspectiveAPI,
  author       = {Google Jigsaw},
  title        = {Perspective API},
  year         = {2025},
  url          = {https://www.perspectiveapi.com/},
  note         = {Accessed: March 16, 2025}
}

@inproceedings{mustafa2024monitoring,
  title={Monitoring The Evolution Of Antisemitic Hate Speech On Extremist Social Media},
  author={Mustafa, Raza Ul and Japkowicz, Nathalie},
  booktitle={2024 IEEE Digital Platforms and Societal Harms (DPSH)},
  pages={1--8},
  year={2024},
  organization={IEEE}
}

@article{mohideen2008language,
  title={The language of Islamophobia in Internet articles},
  author={Mohideen, Haja and Mohideen, Shamimah},
  journal={Intellectual Discourse},
  volume={16},
  number={1},
  year={2008}
}

@inproceedings{aldreabi2023enhancing,
  title={Enhancing automated hate speech detection: Addressing islamophobia and freedom of speech in online discussions},
  author={Aldreabi, Esraa and Blackburn, Jeremy},
  booktitle={Proceedings of the International Conference on Advances in Social Networks Analysis and Mining},
  pages={644--651},
  year={2023}
}

@article{mihalcea2024developments,
  title={How developments in natural language processing help us in understanding human behaviour},
  author={Mihalcea, Rada and Biester, Laura and Boyd, Ryan L and Jin, Zhijing and Perez-Rosas, Veronica and Wilson, Steven and Pennebaker, James W},
  journal={Nature Human Behaviour},
  volume={8},
  number={10},
  pages={1877--1889},
  year={2024},
  publisher={Nature Publishing Group UK London}
}

@article{farrand2023hate,
  title={‘Is This a Hate Speech?’The Difficulty in Combating Radicalisation in Coded Communications on Social media Platforms},
  author={Farrand, Benjamin},
  journal={European Journal on Criminal Policy and Research},
  volume={29},
  number={3},
  pages={477--493},
  year={2023},
  publisher={Springer}
}

@article{jaleel2023islamophobia,
  title={Islamophobia Content Detection Using Natural Language Processing},
  author={Jaleel, Abdul and Anwar, Mehmoon and Ali, Farooq and Mukhtar, Raza and Farooq, Muhammad},
  journal={Journal of Computing \& Biomedical Informatics},
  volume={4},
  number={02},
  pages={88--97},
  year={2023}
}

@inproceedings{kikkisetti2024coded,
  title={Coded Term Discovery for Online Hate Speech Detection},
  author={Kikkisetti, Dhanush and Mustafa, Raza and Melillo, Wendy and Corizzo, Roberto and Boukouvalas, Zois and Gill, Jeff and Japkowicz, Nathalie},
  booktitle={2024 IEEE 11th International Conference on Data Science and Advanced Analytics (DSAA)},
  pages={1--10},
  year={2024},
  organization={IEEE}
}

@article{mozafari2020hate,
  title={Hate speech detection and racial bias mitigation in social media based on BERT model},
  author={Mozafari, Marzieh and Farahbakhsh, Reza and Crespi, No{\"e}l},
  journal={PloS one},
  volume={15},
  number={8},
  pages={e0237861},
  year={2020},
  publisher={Public Library of Science San Francisco, CA USA}
}

@book{howard2017freedom,
  title={Freedom of expression and religious hate speech in Europe},
  author={Howard, Erica},
  year={2017},
  publisher={Routledge}
}

@article{Mustafa2024CanGD,
  title={Can GPT-4 detect subcategories of hatred?},
  author={Raza Ul Mustafa and Noman Ashraf and Nathalie Japkowicz},
  journal={2024 IEEE Digital Platforms and Societal Harms (DPSH)},
  year={2024},
  pages={1-6},
  url={https://api.semanticscholar.org/CorpusID:274642399}
}

@article{Vidgen2019TrajectoriesOI,
  title={Trajectories of Islamophobic hate amongst far right actors on Twitter},
  author={Bertie Vidgen and Taha Yasseri and Helen Z. Margetts},
  journal={ArXiv},
  year={2019},
  volume={abs/1910.05794},
  url={https://api.semanticscholar.org/CorpusID:204509076}
}

@article{Obermaier2021IllBT,
  title={I’ll be there for you? Effects of Islamophobic online hate speech and counter speech on Muslim in-group bystanders’ intention to intervene},
  author={Magdalena Obermaier and Desir{\'e}e Schmuck and Muniba Saleem},
  journal={New Media \& Society},
  year={2021},
  volume={25},
  pages={2339 - 2358},
  url={https://api.semanticscholar.org/CorpusID:239675514}
}

@article{Vidgen2018DetectingWA,
  title={Detecting weak and strong Islamophobic hate speech on social media},
  author={Bertie Vidgen and Taha Yasseri},
  journal={Journal of Information Technology \& Politics},
  year={2018},
  volume={17},
  pages={66 - 78},
  url={https://api.semanticscholar.org/CorpusID:56895481}
}

@article{Weber2025DigitalGC,
  title={Digital Guardians: Can GPT-4, Perspective API, and Moderation API Reliably Detect Hate Speech in Reader Comments of German Online Newspapers?},
  author={Manuel Weber and Moritz Huber and Maximilian Auch and Alexander D{\"o}schl and Max-Emanuel Keller and Peter Mandl},
  journal={2025 38th Conference of Open Innovations Association (FRUCT)},
  year={2025},
  pages={318-326},
  url={https://api.semanticscholar.org/CorpusID:275212795}
}

@article{Saeed2023TopicMB,
  title={Topic Modeling based Text Classification Regarding Islamophobia using Word Embedding and Transformers Techniques},
  author={Ammar Saeed and Hikmat Ullah Khan and Achyut Shankar and Talha Imran and Danish Khan and Muhammad Kamran and Muhammad Attique Khan},
  journal={ACM Transactions on Asian and Low-Resource Language Information Processing},
  year={2023},
  url={https://api.semanticscholar.org/CorpusID:264988694}
}

@inproceedings{Gitari2015ALA,
  title={A Lexicon-based Approach for Hate Speech Detection},
  author={Njagi Dennis Gitari and Zuping Zhang and Zuping Zhang and Hanyurwimfura Damien and Jun Long},
  booktitle={International Conference on Multimedia and Ubiquitous Engineering},
  year={2015},
  url={https://api.semanticscholar.org/CorpusID:16011169}
}

@article{GonzlezBaquero2023TheCA,
  title={The Conversation around Islam on Twitter: Topic Modeling and Sentiment Analysis of Tweets about the Muslim Community in Spain since 2015},
  author={William Gonz{\'a}lez-Baquero and Javier J. Amores and Carlos Arcila-Calder{\'o}n},
  journal={Religions},
  year={2023},
  url={https://api.semanticscholar.org/CorpusID:259007663}
}

@inproceedings{Ahn2024SharedConIH,
  title={SharedCon: Implicit Hate Speech Detection using Shared Semantics},
  author={Hyeseon Ahn and Youngwook Kim and Jungin Kim and Youshin Han},
  booktitle={Annual Meeting of the Association for Computational Linguistics},
  year={2024},
  url={https://api.semanticscholar.org/CorpusID:271909841}
}

@inproceedings{Talat2016HatefulSO,
  title={Hateful Symbols or Hateful People? Predictive Features for Hate Speech Detection on Twitter},
  author={Zeerak Talat and Dirk Hovy},
  booktitle={North American Chapter of the Association for Computational Linguistics},
  year={2016},
  url={https://api.semanticscholar.org/CorpusID:1721388}
}

@article{Bleich2011WhatII,
  title={What Is Islamophobia and How Much Is There? Theorizing and Measuring an Emerging Comparative Concept},
  author={Erik Bleich},
  journal={American Behavioral Scientist},
  year={2011},
  volume={55},
  pages={1581 - 1600},
  url={https://api.semanticscholar.org/CorpusID:143679557}
}

@article{Calderon2021LinguisticPF,
  title={Linguistic Patterns for Code Word Resilient Hate Speech Identification},
  author={Fernando H. Calderon and Namrita Balani and Jherez Taylor and Melvyn Peignon and Yen-Hao Huang and Yi-Shin Chen},
  journal={Sensors (Basel, Switzerland)},
  year={2021},
  volume={21},
  url={https://api.semanticscholar.org/CorpusID:244823905}
}

@inproceedings{Elsherief2018HateLA,
  title={Hate Lingo: A Target-based Linguistic Analysis of Hate Speech in Social Media},
  author={Mai Elsherief and Vivek Kulkarni and Dana Nguyen and William Yang Wang and Elizabeth M. Belding-Royer},
  booktitle={International Conference on Web and Social Media},
  year={2018},
  url={https://api.semanticscholar.org/CorpusID:4809781}
}

\begin{IEEEbiography}[{\raisebox{-0.5\height}{\includegraphics[width=1in,height=1.25in,clip,keepaspectratio]{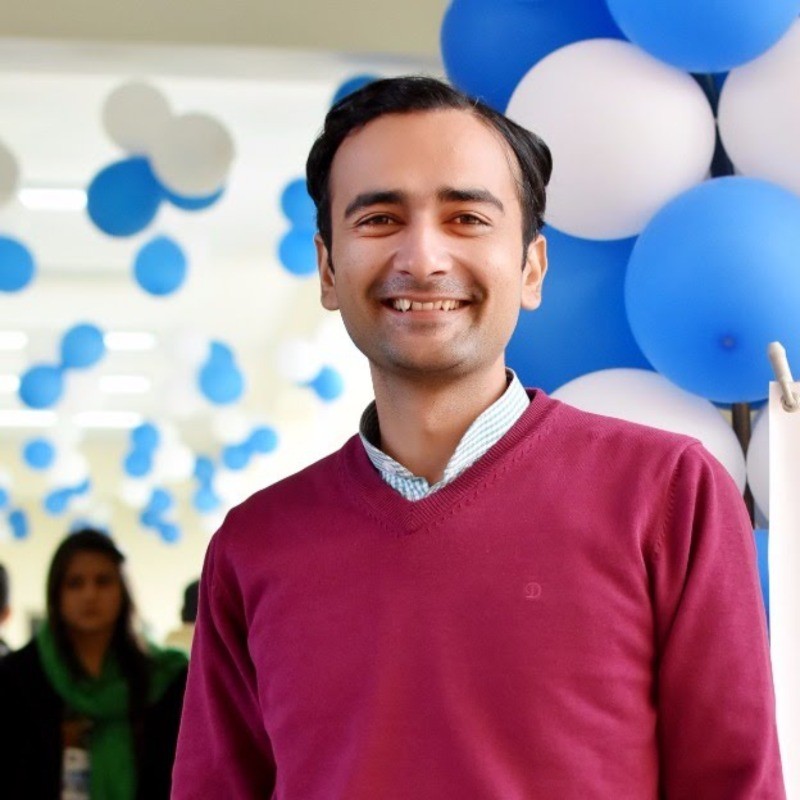}}}]{Raza Ul Mustafa} Ph.D., is an Assistant Professor of Computer Science at Loyola University New Orleans, USA. He received his Ph.D. in Computer Engineering from UNICAMP, Brazil, in 2022 and completed a postdoctoral fellowship in 2024 at American University, USA, focusing on Large Language Models. His research expertise includes Natural Language Processing, social network analysis, and applied machine learning. 

\end{IEEEbiography}

\begin{IEEEbiography}[{\includegraphics[width=1in,height=1.25in,clip,keepaspectratio]{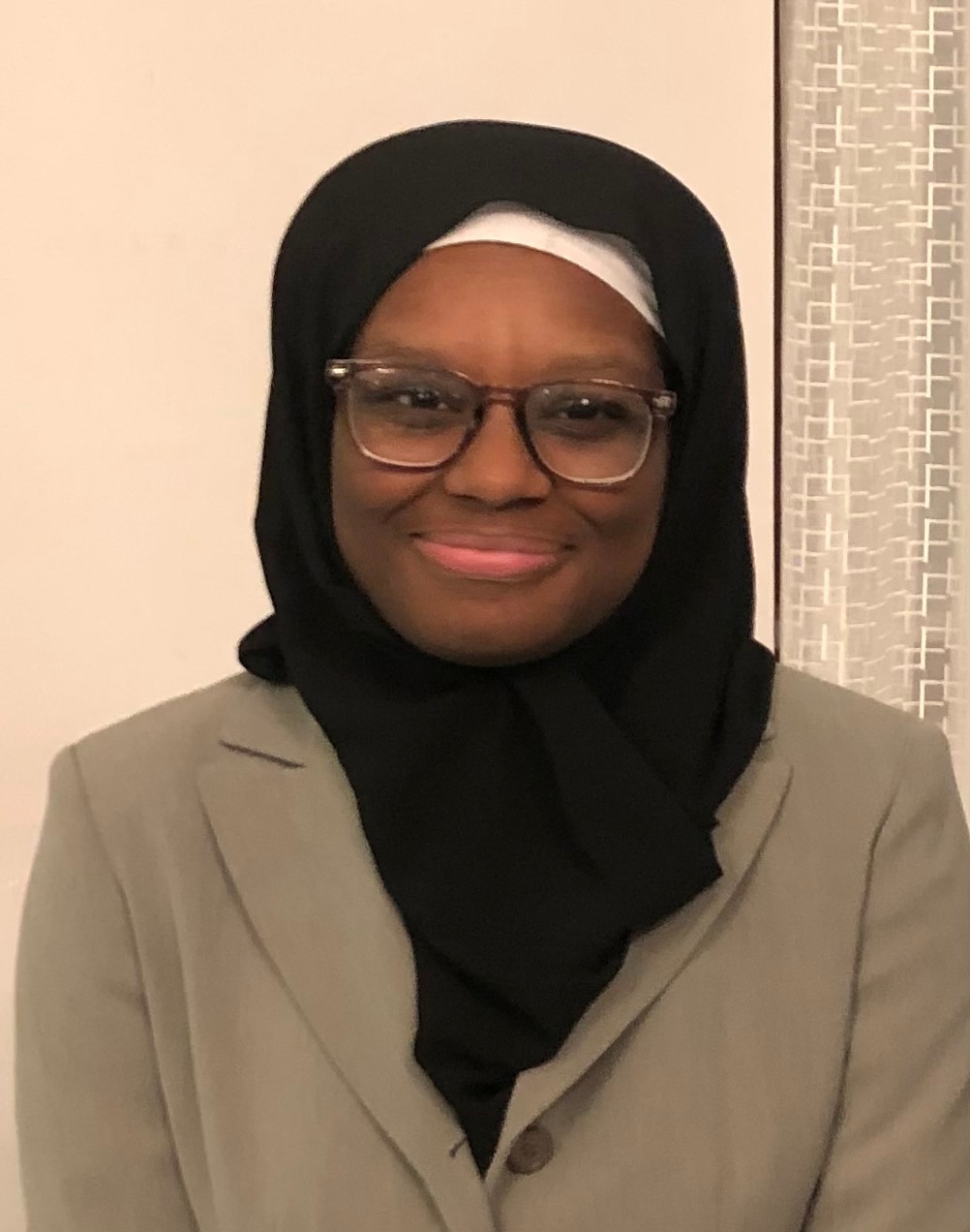}}]{Roi Dupart}
 is a Senior undergraduate student at Loyola University New Orleans studying Computer science. She has conducted machine learning research on GAN architectures, bias modeling in multi-agent medical systems, and LLM-based discourse analysis. Her recent  thesis focused on self-supervised learning for medical data, specifically, representation learning from CT chest scans.
\end{IEEEbiography}

\begin{IEEEbiography}[{\includegraphics[width=1in,height=1.25in,clip,keepaspectratio]{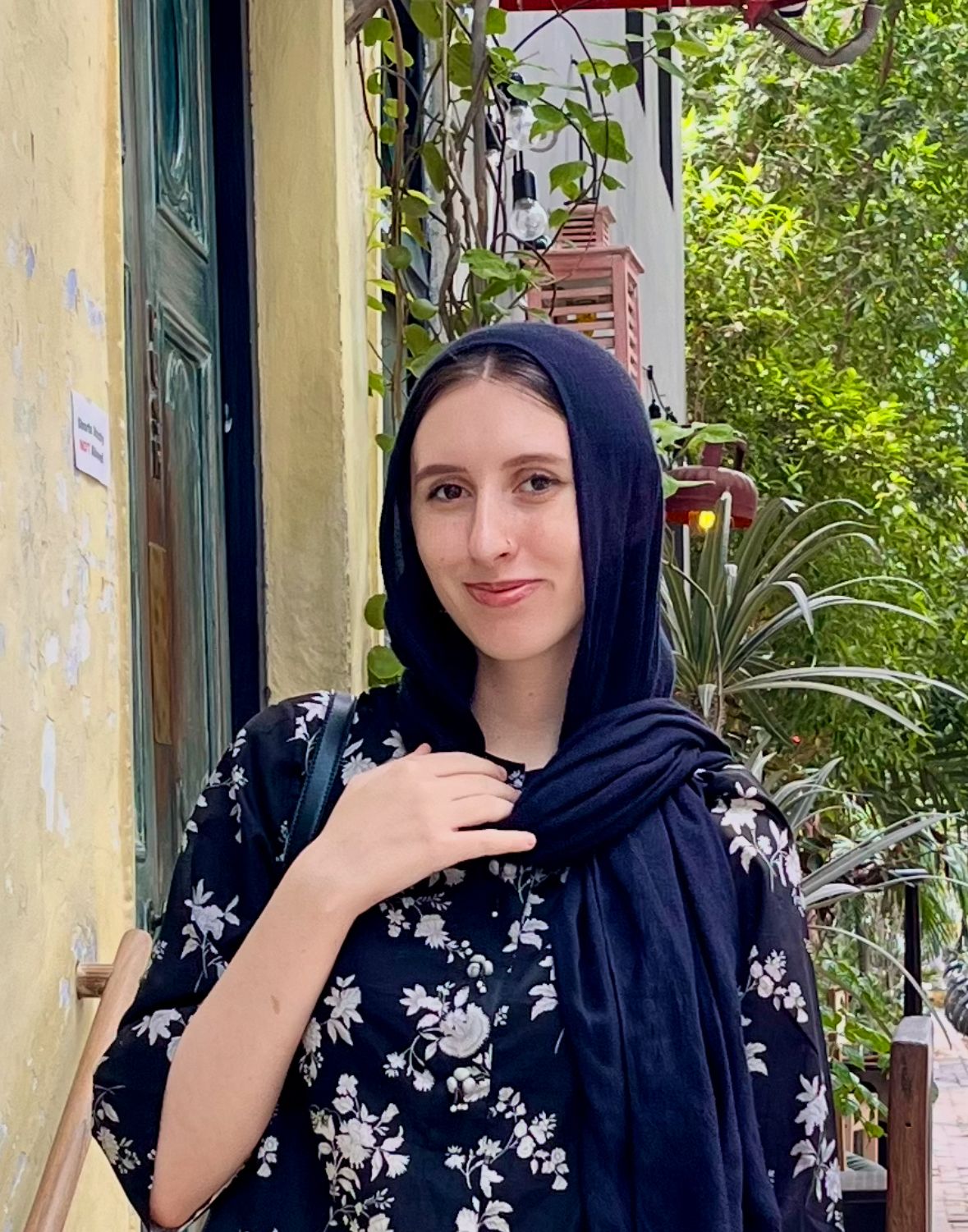}}]{Gabrielle Smith}
Gabrielle Smith is an undergraduate student at Tulane University studying Political Science with a concentration in International Development. Her research interests include South Asia, comparative religion, social-political movements, law, and human rights.
\end{IEEEbiography}

\begin{IEEEbiography}[{\includegraphics[width=1in,height=1.25in,clip,keepaspectratio]{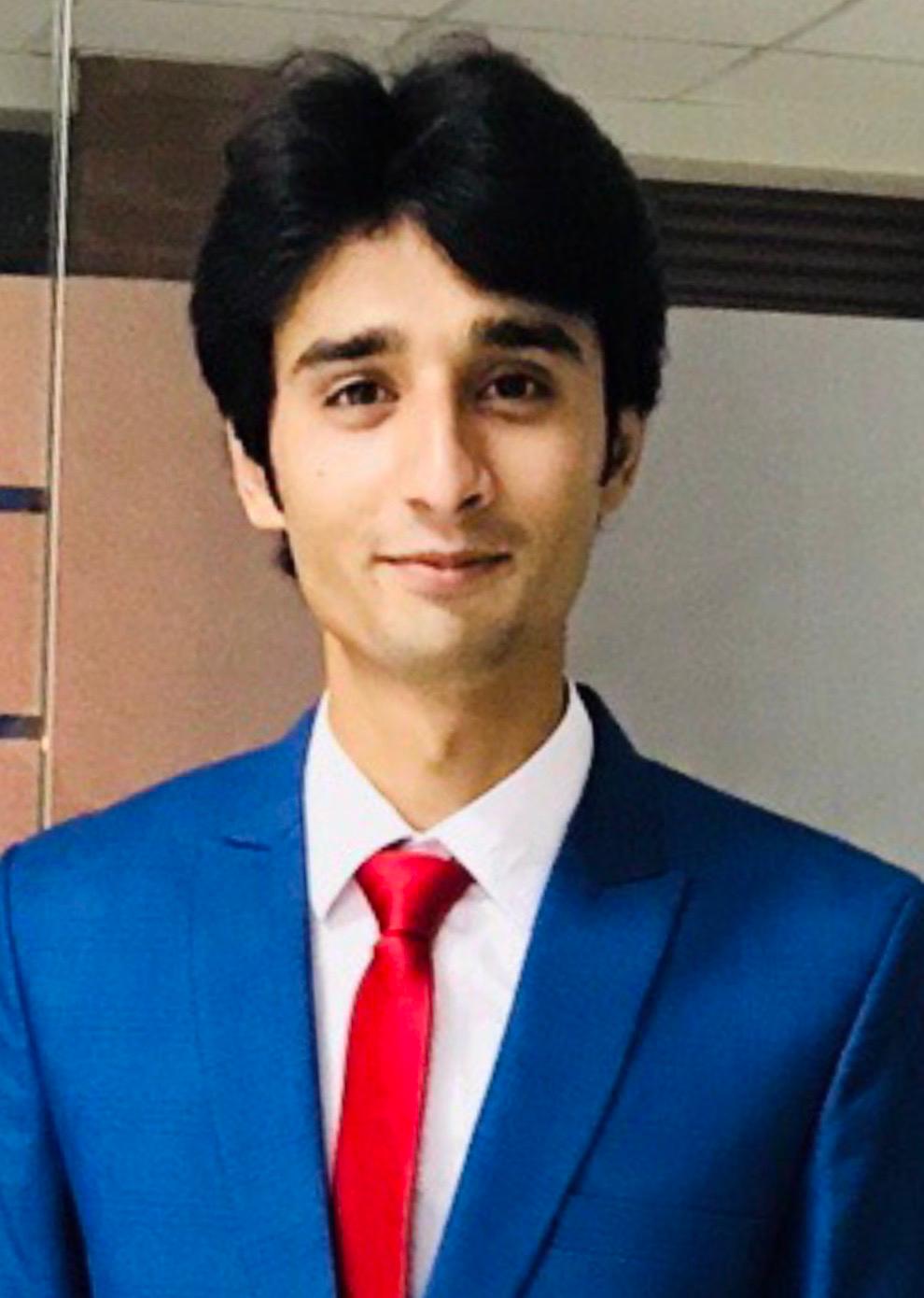}}]{Noman Ashraf} is a Senior Machine Learning Engineer at AbbVie. He holds a PhD in Computer Science from the Instituto Politécnico Nacional – Computing Research Center, with a specialization in Natural Language Processing. Before joining AbbVie, Noman worked as an Associate Scientist at Johnson \& Johnson in Boston, Massachusetts, where he developed AI models for neurodegenerative and neuropsychiatric disorders. He also served as a Postdoctoral Fellow at the Dana-Farber Cancer Institute | Harvard Medical School, focusing on building large-scale AI systems to extract cancer outcomes from electronic health records. Outside of work, he enjoys playing table tennis, volleyball, and cricket.
\end{IEEEbiography}

\begin{IEEEbiography}[{\includegraphics[width=1in,height=1.25in,clip,keepaspectratio]{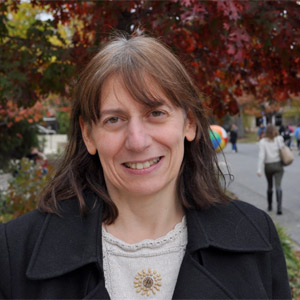}}]{Nathalie Japkowicz} is a professor in the Computer Science Department at American University, which she chaired from July 2018 to June 2024. Prior to that, she directed the Laboratory for Research on Machine Learning applied to Defense and Security at the University of Ottawa in Canada. She is a Professor and AI/Machine Learning researcher particularly interested in lifelong machine learning, anomaly detection, hate speech monitoring, machine learning evaluation, and the handling of uncharacteristic data including datasets plagued by class imbalances.  She trained over 30 graduate students. Her research has been funded by American University’s Signature Research Initiative, DARPA’s L2M program, NSERC, DRDC, Health Canada, and various private companies. Her publications include Evaluating Learning Algorithms: A Classification Perspective at Cambridge University Press (2011), an edited book in the Springer Series on Big Data (2016), and over 150 book chapters, journal articles, and conference or workshop papers. Her recent co-authored book entitled Machine Learning Evaluation: Towards Reliable and Responsible AI at Cambridge University Press appeared in November 2024. She received five best paper awards, including the prestigious European Conference on Machine Learning 2014 Test of Time award, and was awarded the Canadian Artificial Intelligence Association Distinguished Service Award in 2021.
\end{IEEEbiography}
\end{document}